\pdfoutput=1

\documentclass[11pt]{article}

\usepackage[]{template_acl/acl}

\usepackage{times}
\usepackage{latexsym}

\usepackage[T1]{fontenc}

\usepackage[utf8]{inputenc}

\usepackage{microtype}

\usepackage{csquotes}
\usepackage{amsmath}
\usepackage{tabularx,booktabs,multirow}
\usepackage{varwidth}
\usepackage{graphicx}
\usepackage{verbatim}
\usepackage{subcaption}
\usepackage{lscape} 
\usepackage{rotating}
\usepackage{color, colortbl} 
\usepackage{xcolor}

\usepackage{array}
\newcolumntype{+}{>{\global\let\currentrowstyle\relax}}
\newcolumntype{^}{>{\currentrowstyle}}
\newcommand{\rowstyle}[1]{\gdef\currentrowstyle{#1}%
  #1\ignorespaces
}

\definecolor{Gray}{gray}{0.9}

\newcommand{\T}[1]{\textcolor{black}{#1}} 
\newcommand{\M}[1]{\textcolor{black}{#1}} 

\title{What do Toothbrushes do in the Kitchen? \\ How Transformers Think our World is Structured}

\author{Alexander Henlein \and Alexander Mehler \\
         Text Technology Lab, Goethe-University Frankfurt, Germany \\
         \texttt{\{henlein, mehler\}@em.uni-frankfurt.de}}

\begin{document}
\maketitle
\begin{abstract}
Transformer-based models are now predominant in NLP.
They outperform approaches based on static models in many respects.
This success has in turn prompted research that reveals a number of biases in the language models generated by transformers.
In this paper we utilize this research on biases to investigate to what extent transformer-based language models allow \M{for} \T{extracting} knowledge about object relations (\textit{$X$ occurs in $Y$}; \textit{$X$ consists of $Z$}; \textit{action $A$ involves using $X$}).
To this end, we compare contextualized models with their static counterparts. 
We make this comparison dependent on the application of a number of similarity measures and classifiers.
Our results are threefold:
Firstly, we show that the models combined with the different similarity measures differ greatly in terms of the amount of knowledge they allow \M{for} \T{extracting}.
Secondly, our results suggest that similarity measures perform much worse than classifier-based approaches.
Thirdly, we show that, surprisingly, static models perform almost as well as contextualized models -- in some cases even better.
\end{abstract}

\section{Introduction}
Few models have recently influenced NLP as much as transformers~\cite{Vaswani:et:al:2017}.
Hardly any new NLP system today \T{is introduced} without a transformer-based model such as BERT~\cite{Devlin:et:al:2019} or GPT~\cite{Radford:et:al:2019}. 
As a result, static models such as word2vec~\cite{Mikolov:et:al:2013} are increasingly being substituted.
Nevertheless, transformers are still far from being fully understood. 
Thus, research studies are being conducted to find out how they work and what properties the language models they generate have.


\T{During training, transformers seem to capture both syntactic and semantic features~\cite{Rogers:et:al:2020}.}
For example, dependency trees can be reconstructed from trained attention heads~\cite{Clark:et:al:2019}, syntactic trees can be reconstructed from word encodings~\cite{Hewitt:Manning:2019}, and these encodings can be clustered into representations of word senses~\cite{Reif:et:al:2019}.
BERT also seems to encode information about entity types and semantic roles~\cite{Tenney:et:al:2019}.
For an overview of this research see~\newcite{Rogers:et:al:2020}.


Since BERT and other transformers are trained on various data \T{crawled from the internet}, they are sensitive to biases~\cite{Caliskan:et:al:2017,May:et:al:2019,Bender:et:al:2021}.
In practice, instead of reproducing \T{negative} biases, they are expected to allow \M{for} the derivation of statements, such as that toothbrushes are spatially associated with bathrooms rather than living rooms.
In this line of thinking, approaches such as the popularization of knowledge graphs can be located~\T{\cite{Yao:et:al:2019,Petroni:et:al:2019}}.
Our paper is situated in \M{this context}.
More specifically, we examine the extent to which knowledge about spatial objects and their relations is implicitly encoded in these models.
Since the underlying texts are rather implicit regarding such information, it can be assumed that the object relations derivable from transformers are weakly encoded~\cite[cf. ][]{Landau:Jackendoff:1993,Hayward:Tarr:1995}.
\M{Reading,} \T{
for example,} \M{the sentence} \T{
%
\enquote{\textit{After getting up, I ate an apple}}
%
\M{one may assume that the narrator got up from his bed in the bedroom},
went to the kitchen, 
took an apple, washed it in the sink, and finally ate it. 
The apple \M{is also likely to have been} peeled and cut.
\M{Equally, however, nothing is said in the sentence about a bedroom or a kitchen.}}
Nevertheless, it is a well known approach to explore the usage regularities of words, currently most efficiently represented by neural networks, as a source for knowledge extraction~\cite[see, e.g.][]{Zhang:et:al:2017,Bouraoui:et:al:2020,Shin:et:al:2020,Petroni:et:al:2019}. 

In this work, we use a number of methods to identify biases in contextualized models and ask to what extent they can be used to extract object-based knowledge from these models.
To this end, we consider three relations:
\vspace{-0.2cm}
\begin{enumerate}
\setlength\itemsep{0em}
    \item \textit{Spatial containment of (source) objects in (target) rooms:}
    e.g.\ {a fridge probably belongs in a kitchen, but not in a living room};
    \item \textit{Parts (source) in relation to composite objects (target):} 
    e.g.\ {a refrigerator compartment is probably a part of a fridge};
    \item \textit{Objects (source) in relation to actions (target) that involve them:}
    e.g.\ {reading involves something being read, e.g., a book}.
\end{enumerate}
\vspace{-0.2cm}
Regarding these relations, we examine a set of pre-trained contextualized and static word representation models. 
This is done to answer the question to what extent they allow the extraction of instances of these relations when trained on very large datasets.
We focus on rather common terms (\textit{kitchen}, \textit{to read} etc.) as part of the general language.

\T{It is assumed that (static or contextualized) models implicitly represent \M{such} relations, \M{so that it is} possible to identify probable targets starting from certain  sources.}
That is, for a word like \textit{fridge} (source), we expect it to be semantically more strongly associated with \textit{kitchen} (target) than with words naming other rooms, since fridges are more likely to be found in kitchens than in other rooms, and that certain word representation models reflect this association. 
We also assume that this association is asymmetric and exists to a lesser extent from target to source (cf.\ \newcite{Tversky:Gati:2004}).

The paper is organized as follows:
Related work is reported in Sec.\ \ref{Related Work}.
\T{The datasets we use are represented in Sec.\ \ref{sec:Datasets} and our method in Sec.\ \ref{sec:Approach}}.
%
Our experiments are presented in Sec.\ \ref{sec:results} and discussed in Sec.\ \ref{sec:Discussion}. 
%
\M{Sec.\ \ref{sec:Conclusion_Futurework} provides a conclusion.}

\section{Related Work}\label{Related Work}
Biases in NLP models are not a new problem that appeared with BERT, but affect almost all models trained on language datasets~\cite{Caliskan:et:al:2017}.
As such, there are methods for measuring social biases in static models such as word2vec~\cite{Mikolov:et:al:2013}.
One of the best known approaches is WEAT~\cite{Caliskan:et:al:2017}.
Here, two groups of concepts are compared with two groups of attributes based on the difference between the sums of their cosine similarities (see Section \ref{sec:Approach}).
This approach already points to a methodological premise that also guides our work:
Relations of entities are tentatively determined by similarity analyses of vectorial word representations. 

However, a direct comparison of word vectors is not possible with contextualized methods such as BERT, where the vector representation of a word varies with the context of its occurrence\T{~\cite[cf.][]{Ethayarajh:2019}}.
Efforts to transfer the cosine-based approach from static to contextualized models have not been able to recreate similar performances~\cite{May:et:al:2019}.
Therefore, new approaches have been developed based on the specifics of contextualized models.
For example, BERT is trained using masked language modeling, where the model estimates the probability of masked words in sentences~\cite{Devlin:et:al:2019}.
The probability distribution for a masked word in a given context can then be used as information to characterize candidate words~\cite{Kurita:et:al:2019}.
Sec.\ \ref{subsec:Metrics} describes this approach in more detail.
%
%
\T{An \M{alternative} approach \M{is} to examine the interpretability of models~\cite{Belinkov:Glass:2019,Jiang:et:al:2020,Petroni:et:al:2019,Petroni:et:al:2020,Bommasani:et:al:2020,Hupkes:et:al:2020},
\M{which goes} beyond the scope of this paper.
\M{In any event, both approaches share basis basic ideas,} e.g., the probability prediction of mask tokens~\cite[cf.][]{Kurita:et:al:2019,Belinkov:Glass:2019}.
}

Work has also been done on how BERT represents information about spatial objects.
For example, BERT has problems with certain object properties (e.g.\ \textit{cheap} or \textit{cute}) or implicit visual properties that are rarely expressed~\cite{Da:Kasai:2019}.
Problems are also encountered with extracting numerical commonsense knowledge, such as the typical number of tires on a car or the feet on a bird~\cite{Lin:et:al:2020}.
More than that, the models seem to allow \M{for} extracting some object knowledge, but not with respect to properties based on their affordance (e.g.\ objects through which one can see are transparent~\cite{Forbes:et:al:2019}).
Even though these results seem to question the use of BERT and its competitors for knowledge extraction, these models still perform better in downstream tasks than their static competitors \cite{Devlin:et:al:2019,Liu:et:al:2019,Brown:et:al:2020,Da:Kasai:2019}.
\newcite{Bouraoui:et:al:2020} compared these models using different datasets and lexical relations. 
These include relations similar to those examined here (e.g.\ a pot is usually found in a kitchen), but beyond the level of detail achieved in our study.

\T{What will become \M{increasingly} important is the so-called grounding of language models~\cite{Merrill:et:al:2021}:
\M{Here}, the models are trained not only on increasingly large text data, but also, for example, on images thus enabling better \enquote{understanding} of spatial relations~\cite{Sileo:2021,Li:et:al:2020}.
In this paper, we focus on models without grounding.
}

\section{Datasets Used for Evaluation}\label{sec:Datasets}

\subsection{Spatial Containment}
The \textit{NYU Depth V2 Dataset}~\cite{Silberman:et:al:2012} consists of video sequences of numerous indoor scenes.
    It features 464 labeled scenes using a rich category set.
    We use this dataset as a basis for evaluating the probability of occurrence of objects in rooms (e.g.\ kitchen, living room, etc.). 
    That is, we estimate the conditional probability $P(r \mid o)$ of a room $r$ (target) given an object $o$ (source). 
    In this way, we aim to measure the strength of an object's association with a particular room as reflecting the corresponding spatial containment relation. 
    At the same time, we want to filter out objects such as \textit{window} that are evenly distributed among the rooms studied here.
    In our experiments, we consider the ten most frequently mentioned objects in NYU to associate with the five most frequently mentioned spaces.
    This data is shown in the Table \ref{tab:room-obj} (appendix).
    
    The advantage of NYU over other scene datasets such as 3D-Front~\cite{Fu:et:al:2020} is that it deals with real spaces and not artificially created ones.
    In addition, NYU's object category set is relatively fine-grained (we counted 895 different object names) and uses colloquial terms.
    This is in contrast to, for example, SUNCG~\cite{Song:et:al:2017} (with categories like \enquote{slot machine with chair},\enquote{range hood with cabinet}, \enquote{food processor}) and ShapeNetCore~\cite{Chang:et:al:2015} with only 55 object categories or \T{COCO~\cite{Lin:et:al:2014} with 80 object categories.}
    This makes NYU more suitable for our task of evaluating word representation models as resources for knowledge extraction starting from general language.
        
    \subsection{Part-whole Relations}
    We use a subset of the object descriptions from \textit{Online-Bildwörterbuch}\footnote{\url{http://www.bildwoerterbuch.com/en/home}}.
    This resource describes very fine-grained part-whole relations of objects expressed by colloquial names, in contrast to, e.g., PartNet~\cite{Mo:et:al:2019} where one finds labels such as \textit{seat single surface} or \textit{arm near vertical bar}.
    The list of objects from \textit{Online-Bildwörterbuch} used in our study and their subdivisions is shown in Table \ref{tab:obj-part}.
    \T{The selected objects were chosen by hand, provided that the chosen examples are general enough and the subdivision is sufficiently fine.}
    
    \subsection{Action-object Relations}
    To study entities as typical objects of certain actions, we \M{derive} a dataset \M{from} HowToKB~\cite{Chu:et:al:2017} which is based on WikiHow\footnote{\url{https://www.wikihow.com/}}.
    In HowToKB, task frames, temporal sequences of subtasks, and attributes for involved objects were extracted from WikiHow articles.
    \T{Some changes were made to the knowledge database, including a newly crawled version of WikiHow.
    In addition, the pre-processing tools have been updated and partially extended.}
    %
    \M{Our} dataset \M{(see Table~\ref{tab:verb-obj})} will be published on GitHub. 

    \subsubsection{Related Datasets}
    For evaluating static models, there are datasets and approaches to measuring lexical relations, such as DiffVec~\cite{Vylomova:et:al:2015}, BATS~\cite{Gladkova:et:al:2016} \T{or BLiMP~\cite{Warstadt:et:al:2020}}. 
    Although these datasets are also used to evaluate BERT \cite{Bouraoui:et:al:2020}, they represent only an unstructured subset of the data we used and are thus not appropriate for our study.

\section{Approach} \label{sec:Approach}

We \M{now} present the static and contextualized models used in our study.
Table~\ref{tab:model-overview} \T{in the appendix} lists these models and their sources.
We also specify the measures used to compute word associations as a source of knowledge extraction, and describe how to use classifiers as an alternative to them.

\subsection{Static Models}
Probably the best known static model is word2vec \cite{Mikolov:et:al:2013}.
\M{Its CBOW variant} is trained to predict words in the context of their surrounding words.
The word representations trained in this way partially encode semantic relations \cite{Mikolov:et:al:2013}, making them a suitable candidate for comparison with the corresponding information values of contextualized word representations. 
In addition to word2vec, we consider GloVe~\cite{Pennington:et:al:2014}, Levy~\cite{Levy:Goldberg:2014}, fastText~\cite{Mikolov:et:al:2018} \T{and a static BERT model~\cite{Gupta:Jaggi:2021}.}
\T{Unlike window-based approaches \M{to} static embeddings, Levy embeddings are trained on dependency trees.}

\subsection{Contextualized Models}

Unlike static models, the vector representations of (sub-)word (units) in contextualized models depend on the context in which they occur so that tokens of the same type may each be represented differently in different contexts.
All contextual models we evaluate here are pre-trained and come from the \textit{huggingface models repository}\footnote{\url{https://huggingface.co/models}}.
We evaluate two types of contextualized models:

\noindent \textbf{Masked Language Models (MLM)} 
are trained to reconstruct randomly masked words in input sequences.
We experiment with BERT~\cite{Devlin:et:al:2019}, RoBERTa~\cite{Liu:et:al:2019}, ELECTRA~\cite{Clark:et:al:2020} and ALBERT~\cite{Lan:et:al:2019}.
The models differ in training, training data, and model size.
BERT is trained using masked language modeling and next sentence prediction. 
Ro\-BER\-Ta omits the second task, but uses much more training data.
\T{Two models are trained for \M{ELECTRA}: 
one on masked language modeling (generator) and a second \M{one} that recognizes just these replaced tokens (discriminator). 
Since many \M{of our} evaluations need mask tokens, we only use the generator model for the evaluations.}
\M{Finally,} ALBERT is trained to predict the order of pairs of consecutive text segments in addition to masked language modeling.

\noindent \textbf{Causal Language Models (CLM)} are trained to predict the next word for a given input text. 
\M{From} this class we \M{experiment with} 
GPT-2~\cite{Radford:et:al:2019}, GPT-Neo~\cite{Gao:et:al:2020,gpt-neo} and GPT-J~\cite{gpt-j}.
\T{GPT-Neo and GPT-J are re-implementations of GPT-3~\cite{Brown:et:al:2020} where GPT-J was trained on a significantly larger data set named \textit{The Pile}~\cite{Gao:et:al:2021} (cf.\ Table~\ref{tab:model-overview} in the appendix).}

\subsection{Similarity Measures} \label{subsec:Metrics}
To compute similarities of word associations based on the models studied here, we make use of research on biases in such models.
These approaches calculate biases between two groups of concepts with respect to candidate groups of attributes.
To this end, associations are evaluated by computing the similarities of vector representations of concepts and attributes.
We adopt this approach to investigate our research question.
However, as we consider knowledge extraction starting from source words (e.g.\ \textit{toaster}, \textit{shower}) in relation to target words (e.g.\ \textit{kitchen}, \textit{bathroom}), we modify it as described below.

\subsubsection{Cosine and Correlation Measures} \label{subsubsec:Cosine Measure}
Based on the human implicit association test~\cite{Greenwald:et:al:1998}, WEAT \cite{Caliskan:et:al:2017} is originally designed to compare the association between two sets of concepts ($X$ and $Y$) and two sets of attributes ($A$ and $B$).
The degree of bias is calculated as follows:
\begin{align}\footnotesize
s(X, Y, A, B) &= \sum_{x \in X} s(x,A,B) - \sum_{y \in Y} s(y,A,B)\\
s(w, A, B) &= \sum_{a \in A} \cos{(w,a)} - \sum_{b \in B} \cos{(w,b)}
\end{align}
Since we are considering source words in relation to target words, we use the following variant:
\begin{align}
\label{form:cos}
s(X, A) = \frac{1}{|X||A|} \sum_{x \in X} \sum_{a \in A} \cos{(x,a)}
\end{align}
For contextualized models, we adopt the approach of \newcite{May:et:al:2019}, that is, we generate sentences such as \enquote{This is a \{x\}.} or \enquote{A \{x\} is here}.
All templates used in our study are listed in the appendix Table~\ref{tab:context_template}.
%
\T{However, instead of using the BERT token [CLS] (the default token at the beginning of \M{an} input sequence, which often serves as the default representation of the entire sequence), we use the maximum of the vector representations of all subwords of the expression.}
This approach is suitable for models like RoBERTa that do not use the [CLS] token for training, or the GPT models that do not have this token at all.
\T{In addition, we \M{also achieved} slightly better results on regular BERT models using this approach.
We explain this with the fact that our focus is actually only on single tokens and that the vector representation of the [CLS] token often focuses only on a few dimensions~\cite{Zhou:et:al:2019}.}
\M{Our approach} results in a set of contextualized representations for each source and target word, which are then compared using \T{formula ~\ref{form:cos}}.
We were able to obtain better results in our experiments with this representation \T{than with those generated via the [CLS] token.}
For static models, if there is no vector representation for a potential multiword expressions (MWE)\T{\footnote{Word2Vec contains vectors for MWE's.}, the average of the vectors of their components is used.}
\T{This representation yielded the largest Bias in the work of~\newcite{Azarpanah:et:al:2021}.}
For the static models, we also experimented with \textit{distance correlation}~\cite{Szekely:et:al:2007}, \textit{Pearson correlation}~\cite{Benesty:et:al:2009}, \textit{Spearman correlation}~\cite{Kokoska:Zwillinger:2000}, \textit{Kendall's tau}~\cite{Kendall:1938} and \textit{Mahalanobis distance}~\cite{Mahalanobis:1936} -- cf.~\newcite{Torregrossa:et:al:2020,Azarpanah:et:al:2021} \T{\M{of} the word vectors}.
Due to space limitations, only the values of the distance correlation and Kendall's tau are shown (see Table \ref{tab:static-score-summary}); the other correlation measures behave similarly.
Moreover, the values for these measures tend to perform worse for contextualized models.
\T{\M{This observation} is consistent with findings of~\newcite{Azarpanah:et:al:2021} where the {Mahalanobis distance} measure performed worst.} 
%

\subsubsection{Increased Log Probability}
The cosine measure has shown to be problematic for assessing bias in contextualized models such as BERT~\T{\cite{May:et:al:2019,Kurita:et:al:2019}}.
\newcite{Kurita:et:al:2019} have therefore developed a new approach for models trained using masked language modeling.
They weight the probability of a target word in a simple sentence template, assuming that an attribute is given or not:
%
\par\nobreak
\vspace{-0.3cm}
{
\small
\begin{align*}
&\operatorname{score}(\mathit{target}, \mathit{attribute}) = \\
& \log \frac{P([\text{MASK}]=[\mathit{target}]~|~[\text{MASK}]~is~a~[\mathit{attribute}])}
{P([\text{MASK}_{1}]=[\mathit{target}]~|~[\text{MASK}_{1}]~is~a~[\text{MASK}_{2}])}\nonumber
\end{align*}
}%
Experiments show that the values of this measure correlate significantly better with human biases.

Since this measure is based on the context sensitivity of models, it cannot be applied to static models.
For contextualized models, we use the probability of the last token (e.g.\ \textit{curtain} in the case of \textit{shower curtain}) for source-forming MWEs and the first token (e.g.\ \textit{living} in the case of \textit{living room}) for target-forming MWEs.
We also performed experiments with multiple masks, one for each of the components of a MWE.
However, this did not produce better results.
We adapt this approach for causal language models as follows:
Instead of a complete sentence, we use incomplete sentence templates such as \enquote{A(n) \{object\} is usually in the \ldots} or \enquote{In the \{room\} is usually a/an \ldots}. 
The model should then predict the next token.
Instead of masking the seed word, a \T{neutral equivalent is used for calculation}:
\begin{quote}
\centering
    \textit{A(n) \{object\} is usually in the ...} \\
     	$\Downarrow$ \\
    \textit{This is usually in the ....}
\end{quote}
Instead of performing the analysis in only one direction, we determine the score for both the target and the source given the other.

\subsubsection{Classifier-based Measures}
In addition to the \T{previously described} measures, we experiment with classifiers.
To this end, we train three classifiers on the model representations of the source word to determine the associated target word as a class label \T{(e.g. predict \textit{kitchen}, given the vector of \textit{frying pan}).}
We generate the set of source word representations $X$ in the same way as in the case of the cosine measure (see Section \ref{subsubsec:Cosine Measure}) and average them before classification:
\begin{align*}
\mathit{target} = \mathit{Classifier}\left(\frac{1}{|X|}\sum_{\vec{x} \in X} \vec{x}\right)
\end{align*}
The training runs on a leave-one-out cross-va\-li\-da\-tion repeated 100 times.
\T{The target vector was then generated from the counted predicted classes (see Figure~\ref{fig:bert-room-all}b in Appendix)}
We trained a $k$-nearest neighbors classifier with $k=5$ (KNN), an SVM with a linear kernel and a feed-forward network (FFN).
A small hyperparameter optimization was performed for the FFN, which resulted in the following parameters: 
Adam Optimizer~\cite{Kingma:Ba:2014} with a learning rate of ${0.01}$ over ${100}$ epochs and one hidden layer of size ${100}$ \T{and ReLU as activation function}.

\subsection{Scoring Measures and Classifiers}
Given a word representation model, we compute the final score for the measures and classifiers to estimate how well they reconstruct the original probability distribution of the source entities relative to the target entities (see Table \ref{tab:room-obj}, \ref{tab:obj-part}, and \ref{tab:verb-obj}).
This is computed by the distance correlation~\cite{Szekely:et:al:2007} between the target-source probability distributions and the corresponding association distributions of the respective measure or classifier.
The advantage of the distance correlation over the Pearson correlation is that it can also measure nonlinear relations.
This was calculated both for all targets individually (correlation of all sources to one target) and then \textit{concatenated} for all targets together; we denote this variant by \textit{CONC}.
Therefore, \textit{CONC} does not correspond to the average of the individual distance correlations.

\section{Experiments} \label{sec:results}

\begin{table*}
    \setlength{\tabcolsep}{0.5pt}
    \centering
    \tiny
    \begin{tabularx}{\linewidth}{+l ^l | ^X ^X ^X ^X ^X ^X | ^X ^X ^X ^X ^X ^X | ^X ^X ^X ^X ^X ^X | ^X ^X ^X ^X ^X ^X | ^X ^X ^X ^X ^X ^X}
    \toprule
    &  & \multicolumn{6}{c}{\textbf{Word2Vec}} & \multicolumn{6}{c}{\textbf{GloVe}} & \multicolumn{6}{c}{\textbf{Levy}} & \multicolumn{6}{c}{\textbf{fastText}} & \multicolumn{6}{c}{\textbf{static-BERT}}\\ 
    &  & cos & dist & kend & knn & svm & ffn & cos & dist & kend & knn & svm & ffn & cos & dist & kend & knn & svm & ffn & cos & dist & kend & knn & svm & ffn & cos & dist & kend & knn & svm & ffn \\ \midrule
    \multirow{5}{*}{\textbf{\rotatebox[origin=c]{90}{Room}}} 
    & bathroom & 0.37 & 0.37 & 0.37 & 0.39* & 0.62* & 0.82* & 0.38 & 0.39* & 0.38* & 0.57* & 0.93* & 0.93* & 0.39 & 0.40* & 0.39 & 0.14 & 0.34 & 0.37* & 0.53* & 0.53* & 0.52* & 0.73* & 0.67* & 0.90* & 0.54* & 0.50* & 0.50* & 0.25 & 0.66* & 0.70* \\
    
    & bedroom &0.20 & 0.20 & 0.20 & 0.13 & 0.49* & 0.70* & 0.31 & 0.29 & 0.30 & 0.28 & 0.66* & 0.45* & 0.21 & 0.21 & 0.21 & 0.10 & 0.25 & 0.11 & 0.30 & 0.31 & 0.32 & 0.26 & 0.44* & 0.59* & 0.28 & 0.27 & 0.27 & 0.35 & 0.33 & 0.35 \\
    
    & kitchen & 0.35 & 0.34 & 0.35 & 0.20 & 0.55* & 0.53* & 0.37* & 0.40* & 0.41* & 0.52* & 0.65* & 0.81* & 0.17 & 0.17 & 0.18 & 0.09 & 0.32 & 0.30 & 0.38* & 0.36 & 0.34 & 0.41* & 0.66* & 0.76* & 0.40* & 0.41* & 0.41* & 0.45* & 0.53* & 0.68* \\
    
    & living room & 0.23 & 0.23 & 0.24 & 0.06 & 0.33 & 0.35* & 0.30 & 0.27 & 0.28 & 0.10 & 0.49* & 0.51* & 0.24 & 0.24 & 0.23 & 0.40 & 0.16 & 0.25 & 0.25 & 0.26 & 0.24 & 0.09 & 0.36* & 0.60* & 0.19 & 0.19 & 0.19 & 0.00 & 0.10 & 0.46* \\
    
    & office & 0.28 & 0.28 & 0.26 & 0.51* & 0.51* & 0.55* & 0.14 & 0.31 & 0.35 & 0.51* & 0.59* & 0.64* & 0.25 & 0.27 & 0.28 & 0.40 & 0.36 & 0.25 & 0.25 & 0.30 & 0.33 & 0.45* & 0.32 & 0.63* & 0.40* & 0.44* & 0.45* & 0.10 & 0.21 & 0.32 \\ \hline

    \rowcolor{Gray} \rowstyle{\bfseries} & CONC & 0.23* & 0.23* & 0.23* & 0.22* & 0.50* & 0.60* & 0.27* & 0.31* & 0.32* & 0.37* & 0.67* & 0.67* & 0.16 & 0.15 & 0.15 & 0.15 & 0.11 & 0.23* & 0.30* & 0.31* & 0.31* & 0.40* & 0.45* & 0.70* & 0.31* & 0.31* & 0.31* & 0.18* & 0.39* & 0.48* \\ \midrule

    \multirow{6}{*}{\textbf{\rotatebox[origin=c]{90}{Part}}} 
    & bed & 0.41* & 0.41* & 0.40* & 0.64* & 0.56* & 0.56* & 0.38* & 0.51* & 0.51* & 0.56* & 0.76* & 0.84* & - & - & - & - & - & - & 0.42* & 0.51* & 0.52* & 0.69* & 0.61* & 0.67* & 0.47* & 0.48* & 0.46* & 0.16 & 0.59* & 0.54* \\
    
    & dishwasher & 0.19 & 0.23 & 0.23 & 0.06 & 0.37* & 0.27* & 0.33* & 0.32* & 0.30* & 0.03 & 0.19 & 0.32* & - & - & - & - & - & - & 0.35* & 0.33* & 0.33* & 0.06 & 0.13 & 0.23 & 0.17 & 0.17 & 0.17 & 0.13 & 0.28 & 0.31* \\
    
    & door & 0.12 & 0.11 & 0.11 & 0.54* & 0.75* & 0.75* & 0.19 & 0.23 & 0.22 & 0.48* & 0.81* & 0.85* & - & - & - & - & - & - & 0.25 & 0.27 & 0.24 & 0.36* & 0.55* & 0.84* & 0.24 & 0.25 & 0.25 & 0.36* & 0.73* & 0.67* \\
    
    & mortise lock & 0.15 & 0.16 & 0.16 & 0.16 & 0.50* & 0.54* & 0.22 & 0.26 & 0.28* & 0.45* & 0.74* & 0.68* & - & - & - & - & - & - & 0.11 & 0.17 & 0.20 & 0.68* & 0.55* & 0.68* & 0.20 & 0.21 & 0.21 & 0.14 & 0.49* & 0.47* \\

    & refrigerator & 0.44* & 0.46* & 0.46* & 0.51* & 0.47* & 0.52* & 0.53* & 0.57* & 0.56* & 0.55* & 0.55* & 0.66* & - & - & - & - & - & - & 0.54* & 0.58* & 0.58* & 0.28* & 0.40* & 0.55* & 0.50* & 0.50* & 0.50* & 0.56* & 0.56* & 0.53* \\
    
    & toilet & 0.28 & 0.28 & 0.28 & 0.01 & 0.49* & 0.55* & 0.33* & 0.33* & 0.32* & 0.31* & 0.63* & 0.60* & - & - & - & - & - & - & 0.37* & 0.34* & 0.33* & 0.55* & 0.50* & 0.72* & 0.24 & 0.23 & 0.23 & 0.34* & 0.57* & 0.58* \\ \hline

    \rowcolor{Gray} \rowstyle{\bfseries} & CONC & 0.25* & 0.27* & 0.26* & 0.28* & 0.52* & 0.53* & 0.30* & 0.34* & 0.34* & 0.39* & 0.60* & 0.65* & - & - & - & - & - & - & 0.28* & 0.33* & 0.33* & 0.35* & 0.43* & 0.61* & 0.29* & 0.29* & 0.29* & 0.23* & 0.54* & 0.52* \\ \midrule
    
    \multirow{6}{*}{\textbf{\rotatebox[origin=c]{90}{Verb}}} 
    & eat & 0.79* & 0.79* & 0.77* & 0.89* & 0.89* & 0.89* & 0.77* & 0.86* & 0.80* & 0.89* & 0.89* & 0.92* & 0.46* & 0.45* & 0.45* & 0.66* & 0.87* & 0.87* & 0.73* & 0.80* & 0.79* & 0.69* & 0.89* & 0.89* & 0.83* & 0.84* & 0.83* & 0.61 & 0.89* & 0.87*
 \\
    
    & listen to & 0.54* & 0.64* & 0.56* & 0.21 & 0.38* & 0.46* & 0.59* & 0.70* & 0.65* & 0.06 & 0.53* & 0.49* & 0.28 & 0.22 & 0.23 & 0.20 & 0.38* & 0.52* & 0.42* & 0.53* & 0.63* & 0.21 & 0.42* & 0.40* & 0.54* & 0.56* & 0.53* & 0.00 & 0.39* & 0.50* \\
    
    & play & 0.64* & 0.69* & 0.64* & 0.60* & 0.66* & 0.60* & 0.65* & 0.80* & 0.73* & 0.43 & 0.45* & 0.45* & 0.44* & 0.45* & 0.43* & 0.41 & 0.50* & 0.57* & 0.63* & 0.69* & 0.68* & 0.28 & 0.66* & 0.66* & 0.56* & 0.56* & 0.54* & 0.00 & 0.49* & 0.63*
 \\
    
    & read & 0.43* & 0.52* & 0.48* & 0.38* & 0.59* & 0.61* & 0.51* & 0.60* & 0.59* & 0.48* & 0.53* & 0.50* & 0.31 & 0.31 & 0.31 & 0.49* & 0.31 & 0.50* & 0.54* & 0.56* & 0.59* & 0.42* & 0.50* & 0.59* & 0.48* & 0.52* & 0.48* & 0.00 & 0.31 & 0.47* \\
    
    & wash with & 0.53* & 0.54* & 0.53* & 0.48* & 0.61* & 0.63* & 0.48* & 0.57* & 0.53* & 0.66* & 0.66* & 0.62* & 0.37 & 0.34 & 0.35 & 0.41* & 0.66* & 0.62* & 0.45* & 0.51* & 0.49* & 0.67* & 0.66* & 0.66* & 0.39* & 0.40* & 0.40* & 0.11 & 0.55* & 0.61* \\
    
    & wear & 0.76* & 0.78* & 0.76* & 0.88* & 0.84* & 0.88* & 0.80* & 0.87* & 0.84* & 0.88* & 0.83* & 0.85* & 0.56* & 0.52* & 0.50* & 0.82* & 0.85* & 0.85* & 0.77* & 0.80* & 0.79* & 0.59* & 0.93* & 0.92* & 0.78* & 0.82* & 0.80* & 0.72* & 0.81* & 0.84* \\ \hline

    \rowcolor{Gray} \rowstyle{\bfseries} & CONC & 0.58* & 0.60* & 0.57* & 0.56* & 0.64* & 0.67* & 0.59* & 0.68* & 0.65* & 0.55* & 0.65* & 0.65* & 0.34* & 0.32* & 0.31* & 0.46* & 0.59* & 0.65* & 0.51* & 0.58* & 0.58* & 0.43* & 0.66* & 0.68* & 0.54* & 0.55* & 0.54* & 0.15* & 0.56* & 0.65* \\
    \bottomrule
    \end{tabularx}
    \caption{All results of the static models. cos: Cosine Measure, dist: Distance Correlation, kend: Kendall's Tau, knn: K-Nearest Neighbors, svm: Support Vector Machine, fnn: Feed-Forward Network. The gap in Levy is due to its small training set and the corresponding small vocabulary. \T{(* indicates significant \M{values} at $p < 0.01$)}}
    \label{tab:static-score-summary}
\end{table*}

\begin{table*}
    \centering
    \setlength{\tabcolsep}{0.5pt}
    
    \tiny
    \begin{tabularx}{\linewidth}{+l ^l | ^X ^X ^X ^X ^X ^X | ^X ^X ^X ^X ^X ^X | ^X ^X ^X ^X ^X ^X | ^X ^X ^X ^X ^X ^X | ^X ^X ^X ^X ^X ^X}
    \toprule
    &  & \multicolumn{6}{c}{\textbf{BERT-Base}} & \multicolumn{6}{c}{\textbf{BERT-Large}} & \multicolumn{6}{c}{\textbf{RoBERTa}} & \multicolumn{6}{c}{\textbf{ElectraGen}} & \multicolumn{6}{c}{\textbf{Albert}} \\ 
    &  & cos & m-s & m-t & knn & svm & ffn & cos & m-s & m-t & knn & svm & ffn & cos & m-s & m-t & knn & svm & ffn & cos & m-s & m-t & knn & svm & ffn & cos & m-s & m-t & knn & svm & ffn \\ \midrule
    
    \multirow{5}{*}{\textbf{\rotatebox[origin=c]{90}{Room}}} 
    & bathroom & 0.57* & 0.13 & 0.52* & 0.72* & 0.87* & 0.93* & 0.65* & 0.30 & 0.59* & 0.78* & 0.93* & 0.93* & 0.21 & 0.24 & 0.52* & 0.55* & 0.83* & 0.88* & 0.58* & 0.32 & 0.34 & 0.49* & 0.72* & 0.73* & 0.24 & 0.18 & 0.39 & 0.52* & 0.75* & 0.90* \\
    
    & bedroom & 0.48* & 0.33 & 0.43* & 0.53* & 0.66* & 0.77* & 0.44* & 0.41* & 0.44* & 0.44* & 0.87* & 0.78* & 0.23 & 0.18 & 0.36 & 0.17 & 0.53* & 0.60* & 0.32 & 0.31 & 0.37 & 0.37 & 0.37 & 0.39* & 0.23 & 0.22 & 0.47* & 0.31 & 0.44* & 0.68* \\
    
    & kitchen & 0.56* & 0.25 & 0.58* & 0.62* & 0.81* & 0.83* & 0.43* & 0.24 & 0.54* & 0.72* & 0.77* & 0.79* & 0.39 & 0.27 & 0.59* & 0.16 & 0.62* & 0.73* & 0.34 & 0.24 & 0.36 & 0.48* & 0.34 & 0.39* & 0.25 & 0.17 & 0.30 & 0.05 & 0.56* & 0.69* \\
    
    & living room & 0.30 & 0.37 & 0.26 & 0.51* & 0.78* & 0.79* & 0.23 & 0.38 & 0.24 & 0.57* & 0.49* & 0.66* & 0.13 & 0.38 & 0.28 & 0.49* & 0.74* & 0.65* & 0.26 & 0.48* & 0.33 & 0.15 & 0.27 & 0.26 & 0.15 & 0.35 & 0.54* & 0.20 & 0.29 & 0.40* \\
    
    & office & 0.46* & 0.39* & 0.28 & 0.40* & 0.59* & 0.61* & 0.40* & 0.37 & 0.31 & 0.25 & 0.52* & 0.71* & 0.14 & 0.37* & 0.38 & 0.18 & 0.74* & 0.63* & 0.17 & 0.37* & 0.23 & 0.42 & 0.27 & 0.36 & 0.23 & 0.22 & 0.42* & 0.45* & 0.66* & 0.81* \\ \hline
    
    \rowcolor{Gray} \rowstyle{\bfseries} & CONC & 0.43* & 0.26* & 0.33* & 0.54* & 0.73* & 0.78* & 0.34* & 0.26* & 0.36* & 0.55* & 0.72* & 0.78* & 0.19* & 0.22* & 0.31* & 0.28* & 0.69* & 0.71* & 0.22* & 0.30* & 0.27* & 0.38* & 0.40* & 0.43* & 0.19* & 0.15 & 0.23* & 0.25* & 0.53* & 0.69* \\ \midrule
    
    \multirow{6}{*}{\textbf{\rotatebox[origin=c]{90}{Part}}} 
    & bed & 0.55* & 0.41* & 0.51* & 0.51* & 0.69* & 0.79* & 0.49* & 0.41* & 0.55* & 0.56* & 0.69* & 0.69* & 0.20 & 0.42* & 0.62* & 0.49* & 0.52* & 0.60* & 0.37* & 0.31* & 0.43* & 0.44* & 0.44* & 0.43* & 0.26 & 0.40* & 0.54* & 0.36* & 0.66* & 0.71* \\
    
    & dishwasher & 0.22 & 0.16 & 0.22 & 0.27 & 0.31* & 0.28* & 0.30* & 0.18 & 0.31* & 0.29* & 0.17 & 0.18 & 0.16 & 0.19 & 0.19 & 0.13 & 0.24 & 0.17 & 0.26 & 0.19 & 0.21 & 0.01 & 0.23 & 0.36* & 0.17 & 0.18 & 0.25 & 0.26 & 0.25 & 0.23 \\
    
    & door & 0.19 & 0.32* & 0.20 & 0.34* & 0.65* & 0.63* & 0.13 & 0.28 & 0.39* & 0.47* & 0.60* & 0.62* & 0.15 & 0.33* & 0.27 & 0.52* & 0.42* & 0.51* & 0.14 & 0.20 & 0.17 & 0.41* & 0.57* & 0.60* & 0.13 & 0.29* & 0.21 & 0.36* & 0.50* & 0.54* \\
    
    & mortise lock & 0.12 & 0.14 & 0.09 & 0.16 & 0.26 & 0.28* & 0.14 & 0.23 & 0.11 & 0.19 & 0.26 & 0.35* & 0.07 & 0.29* & 0.12 & 0.08 & 0.18 & 0.28 & 0.16 & 0.18 & 0.15 & 0.39 & 0.59* & 0.39* & 0.09 & 0.27* & 0.22 & 0.16 & 0.31* & 0.39* \\
    
    & refrigerator & 0.44* & 0.21 & 0.40* & 0.48* & 0.47* & 0.54* & 0.38* & 0.21 & 0.54* & 0.42* & 0.51* & 0.50* & 0.18 & 0.38* & 0.45* & 0.49* & 0.43* & 0.49* & 0.37* & 0.33* & 0.43* & 0.46* & 0.45* & 0.53* & 0.44* & 0.27* & 0.51* & 0.66* & 0.51* & 0.61* \\
    
    & toilet & 0.18 & 0.16 & 0.29* & 0.16 & 0.34* & 0.45* & 0.25 & 0.16 & 0.26 & 0.36* & 0.55* & 0.50* & 0.22 & 0.34* & 0.41* & 0.45* & 0.51* & 0.51* & 0.34* & 0.26 & 0.42* & 0.26 & 0.41* & 0.46* & 0.24 & 0.23 & 0.25 & 0.22 & 0.31* & 0.46* \\ \hline
    
    \rowcolor{Gray} \rowstyle{\bfseries} & CONC & 0.20* & 0.20* & 0.24* & 0.33* & 0.45* & 0.49* & 0.22* & 0.21* & 0.28* & 0.39* & 0.46* & 0.46* & 0.07 & 0.29* & 0.29* & 0.39* & 0.39* & 0.43* & 0.21* & 0.19* & 0.23* & 0.32* & 0.45* & 0.47* & 0.08 & 0.23* & 0.27* & 0.35* & 0.42* & 0.49* \\ \midrule

    \multirow{6}{*}{\textbf{\rotatebox[origin=c]{90}{Verb}}} 
    & eat & 0.78* & 0.65* & 0.67* & 0.89* & 0.84* & 0.90* & 0.65* & 0.58* & 0.72* & 0.80* & 0.89* & 0.90* & 0.26 & 0.66* & 0.81* & 0.65* & 0.87* & 0.86* & 0.62* & 0.64* & 0.76* & 0.74* & 0.79* & 0.79* & 0.53* & 0.61* & 0.74* & 0.57* & 0.84* & 0.85* \\
    
    & listen to & 0.46* & 0.53* & 0.51* & 0.42* & 0.52* & 0.57* & 0.50* & 0.52* & 0.50* & 0.43* & 0.55* & 0.52* & 0.30 & 0.53* & 0.55* & 0.23 & 0.49* & 0.54* & 0.57* & 0.47* & 0.59* & 0.00 & 0.36* & 0.39* & 0.23 & 0.47* & 0.51* & 0.07 & 0.44* & 0.57* \\
    
    & play & 0.63* & 0.58* & 0.69* & 0.54* & 0.58* & 0.61* & 0.55* & 0.60* & 0.73* & 0.54* & 0.64* & 0.66* & 0.37 & 0.64* & 0.65* & 0.38* & 0.53* & 0.59* & 0.64* & 0.53* & 0.69* & 0.64* & 0.64* & 0.65* & 0.37 & 0.42* & 0.52* & 0.45 & 0.60* & 0.62* \\
    
    & read & 0.42* & 0.46* & 0.65* & 0.34 & 0.73* & 0.65* & 0.30 & 0.42* & 0.66* & 0.42* & 0.77* & 0.59* & 0.26 & 0.29 & 0.59* & 0.21 & 0.44* & 0.44* & 0.41* & 0.43* & 0.63* & 0.51* & 0.68* & 0.69* & 0.31 & 0.19 & 0.57* & 0.35 & 0.63* & 0.60* \\
    
    & wash with & 0.49* & 0.46* & 0.33* & 0.49* & 0.66* & 0.63* & 0.42* & 0.53* & 0.45* & 0.61* & 0.62* & 0.60* & 0.30 & 0.56* & 0.30 & 0.23 & 0.60* & 0.59* & 0.42* & 0.50* & 0.35* & 0.52* & 0.40 & 0.41 & 0.33 & 0.42* & 0.32 & 0.18 & 0.46* & 0.51* \\

    & wear & 0.66* & 0.64* & 0.76* & 0.88* & 0.90* & 0.92* & 0.62* & 0.57* & 0.74* & 0.79* & 0.90* & 0.85* & 0.24 & 0.64* & 0.77* & 0.36* & 0.72* & 0.79* & 0.53* & 0.62* & 0.74* & 0.90* & 0.84* & 0.83* & 0.30 & 0.61* & 0.77* & 0.61* & 0.77* & 0.86*  \\ \hline
    
    \rowcolor{Gray} \rowstyle{\bfseries} & CONC & 0.53* & 0.53* & 0.38* & 0.59* & 0.69* & 0.71* & 0.37* & 0.50* & 0.44* & 0.60* & 0.73* & 0.68* & 0.20* & 0.55* & 0.37* & 0.28* & 0.59* & 0.64* & 0.49* & 0.51* & 0.37* & 0.60* & 0.61* & 0.62* & 0.15* & 0.40* & 0.26* & 0.29* & 0.62* & 0.67* \\
    
    \bottomrule
    \end{tabularx}
    \caption{All results of the contextual masked-language models. cos: Cosine Measure, m-s: Masked-Source Log Score, m-t: Masked-Target Log Score, knn: K-Nearest Neighbors, svm: Support Vector Machine, fnn: Feed-Forward Network. \T{(* indicates significant \M{values} at $p < 0.01$)}}
    \label{tab:masked-score-summary}
\end{table*}

\begin{table*}
    \centering
    \setlength{\tabcolsep}{2pt}
    \tiny
    \begin{tabularx}{\linewidth}{+l ^l | ^X ^X ^X ^X ^X ^X ^X ^X | ^X ^X ^X ^X ^X ^X ^X ^X | ^X ^X ^X ^X ^X ^X ^X ^X }
    \toprule
    &  & \multicolumn{8}{c}{\textbf{GPT2}} & \multicolumn{8}{c}{\textbf{GPT-Neo}} & \multicolumn{8}{c}{\textbf{GPT-J}} \\ 
    &  & cos & p-s & p-s-l & p-t & p-t-l & knn & svm & ffn & cos & p-s & p-s-l & p-t & p-t-l & knn & svm & ffn & cos & p-s & p-s-l & p-t & p-t-l & knn & svm & ffn  \\ \midrule
    
    \multirow{5}{*}{\textbf{\rotatebox[origin=c]{90}{Room}}}
    & bathroom & 0.52* & 0.20 & 0.38* & 0.50* & 0.37* & 0.31 & 0.95* & 0.95* & 0.30 & 0.22 & 0.51* & 0.36 & 0.25 & 0.53* & 0.89* & 0.91* & 0.50* & 0.26 & 0.60* & 0.66* & 0.48* & 0.35 & 0.89* & 0.92* \\
    
    & bedroom & 0.26 & 0.31 & 0.23 & 0.47* & 0.38* & 0.26 & 0.61* & 0.54* & 0.19 & 0.33 & 0.21 & 0.53* & 0.48* & 0.55* & 0.49* & 0.57* & 0.24 & 0.32 & 0.23 & 0.62* & 0.48* & 0.33 & 0.70* & 0.64* \\
    
    & kitchen & 0.34 & 0.41* & 0.45* & 0.69* & 0.60* & 0.53* & 0.82* & 0.83* & 0.31 & 0.49* & 0.67* & 0.70* & 0.57* & 0.38* & 0.51* & 0.81* & 0.33 & 0.36* & 0.52* & 0.83* & 0.70* & 0.70* & 0.82* & 0.83* \\
    
    & living room & 0.21 & 0.43* & 0.26 & 0.41* & 0.33 & 0.16 & 0.27 & 0.46* & 0.26 & 0.57* & 0.39 & 0.60* & 0.44* & 0.28 & 0.13 & 0.48* & 0.21 & 0.50* & 0.47* & 0.67* & 0.45* & 0.46 & 0.40* & 0.63* \\
    
    & office & 0.13 & 0.21 & 0.43* & 0.37 & 0.23 & 0.31 & 0.44* & 0.73* & 0.33 & 0.30 & 0.43* & 0.46* & 0.34 & 0.24 & 0.53* & 0.72* & 0.23 & 0.36 & 0.53* & 0.49* & 0.39 & 0.37* & 0.52* & 0.69* \\ \hline
    
    \rowcolor{Gray} \rowstyle{\bfseries} & CONC & 0.26* & 0.23* & 0.30* & 0.46* & 0.35* & 0.30* & 0.61* & 0.72* & 0.15 & 0.34* & 0.44* & 0.44* & 0.35* & 0.40* & 0.52* & 0.71* & 0.23* & 0.32* & 0.42* & 0.56* & 0.41* & 0.42* & 0.66* & 0.74* \\ \midrule
    
    \multirow{6}{*}{\textbf{\rotatebox[origin=c]{90}{Part}}}
    & bed & 0.36* & 0.30 & 0.51* & 0.67* & 0.45* & 0.55* & 0.59* & 0.70* & 0.32* & 0.38* & 0.46* & 0.77* & 0.70* & 0.66* & 0.78* & 0.88* & 0.46* & 0.38* & 0.36* & 0.81* & 0.68* & 0.71* & 0.83* & 0.84* \\
    
    & dishwasher & 0.11 & 0.34* & 0.22 & 0.25 & 0.23 & 0.18 & 0.21 & 0.28* & 0.06 & 0.23 & 0.30* & 0.30* & 0.29* & 0.15 & 0.15 & 0.24 & 0.09 & 0.26 & 0.30* & 0.44* & 0.38* & 0.12 & 0.15 & 0.32* \\
    
    & door & 0.23 & 0.07 & 0.14 & 0.20 & 0.28 & 0.20 & 0.65* & 0.66* & 0.27 & 0.10 & 0.17 & 0.35* & 0.42* & 0.20 & 0.44* & 0.66* & 0.15 & 0.13 & 0.12 & 0.37* & 0.41* & 0.25 & 0.67* & 0.77* \\
    
    & mortise lock & 0.07 & 0.34* & 0.43* & 0.17 & 0.18 & 0.27 & 0.63* & 0.65* & 0.11 & 0.49* & 0.42* & 0.30 & 0.22 & 0.27 & 0.49* & 0.61* & 0.15 & 0.43* & 0.43* & 0.47* & 0.31* & 0.04 & 0.63* & 0.66* \\
    
    & refrigerator & 0.42* & 0.41* & 0.24 & 0.53* & 0.52* & 0.47* & 0.39* & 0.51* & 0.29* & 0.33* & 0.33* & 0.47* & 0.55* & 0.44* & 0.47* & 0.57* & 0.46* & 0.51* & 0.41* & 0.57* & 0.63* & 0.55* & 0.53* & 0.63* \\
    
    & toilet & 0.29* & 0.36* & 0.44* & 0.20 & 0.17 & 0.16 & 0.49* & 0.54* & 0.27 & 0.42* & 0.50* & 0.25 & 0.26 & 0.23 & 0.50* & 0.58* & 0.32* & 0.45* & 0.48* & 0.32* & 0.37* & 0.26 & 0.53* & 0.62* \\ \hline
    
    \rowcolor{Gray} \rowstyle{\bfseries} & CONC & 0.14* & 0.24* & 0.28* & 0.28* & 0.25* & 0.29* & 0.47* & 0.54* & 0.12 & 0.30* & 0.34* & 0.37* & 0.36* & 0.32* & 0.46* & 0.57* & 0.16* & 0.34* & 0.32* & 0.42* & 0.43* & 0.33* & 0.54* & 0.62* \\ \midrule
    
    \multirow{6}{*}{\textbf{\rotatebox[origin=c]{90}{Verb}}}
    & eat & 0.49* & 0.82* & 0.65* & - & - & 0.82* & 0.87* & 0.87* & 0.45* & 0.86* & 0.66* & - & - & 0.76* & 0.87* & 0.88* & 0.48* & 0.68* & 0.74* & - & - & 0.63* & 0.89* & 0.89* \\
    
    & listen to & 0.22 & 0.57* & 0.51* & - & - & 0.29 & 0.50* & 0.58* & 0.22 & 0.47 & 0.47* & - & - & 0.20 & 0.42* & 0.55* & 0.21 & 0.60* & 0.56* & - & - & 0.29 & 0.52* & 0.60* \\
    
    & play & 0.40* & 0.64* & 0.62* & - & - & 0.62* & 0.61* & 0.59* & 0.32 & 0.66* & 0.61* & - & - & 0.20 & 0.55* & 0.62* & 0.34 & 0.66* & 0.70* & - & - & 0.37 & 0.67* & 0.67* \\
    
    & read & 0.32 & 0.63* & 0.30 & - & - & 0.32* & 0.45* & 0.45* & 0.32 & 0.61* & 0.34 & - & - & 0.29 & 0.52* & 0.49* & 0.40* & 0.63* & 0.41* & - & - & 0.20 & 0.59* & 0.49* \\
    
    & Wash with & 0.39* & 0.77* & 0.51* & - & - & 0.57* & 0.61* & 0.63* & 0.28 & 0.66* & 0.52* & - & - & 0.39 & 0.41 & 0.60* & 0.23 & 0.69* & 0.52* & - & - & 0.66* & 0.61* & 0.64* \\
    
    & wear & 0.44* & 0.38 & 0.72* & - & - & 0.76* & 0.84* & 0.87* & 0.16 & 0.39 & 0.66* & - & - & 0.68* & 0.79* & 0.85* & 0.21 & 0.38 & 0.62* & - & - & 0.87* & 0.84* & 0.87* \\ \hline
    
    \rowcolor{Gray} \rowstyle{\bfseries} & CONC & 0.31* & 0.52* & 0.53* & - & - & 0.51* & 0.64* & 0.66* & 0.19* & 0.49* & 0.52* & - & - & 0.40* & 0.59* & 0.66* & 0.23* & 0.50* & 0.56* & - & - & 0.48* & 0.68* & 0.69* \\

    \bottomrule
    \end{tabularx}
    \caption{All results of the contextual causal-language models. p-s: Predict Source Score, p-s-l: : Predict Source Log Score, p-t: Predict Target Score, p-t-l: Predict Target Log Score. The gap for the verb p-t score is due to the lack of an easily applicable sentence templates in this direction. \T{(* indicates significant \M{values} at $p < 0.01$)}}
    \label{tab:gpt-score-summary}
\end{table*}

Using the apparatus of Section \ref{sec:Approach}, we now evaluate the classes of word representation models (static, MLMs and CLMs) in conjunction with the similarity measures and classifiers.
The results for the static models are shown in Table~\ref{tab:static-score-summary}, for the MLMs in Table~\ref{tab:masked-score-summary} and for the CLMs in Table~\ref{tab:gpt-score-summary}.
\T{Figure \ref{fig:bert-room-all}, \ref{fig:bert-part-all} and \ref{fig:bert-verb-all} in Appendix show a visualization of the associations computed by \M{means of} {cosine, masked-target \& masked-source increased log} similarity measures and the FFN classifier based on BERT-Large using the different datasets.}
%

\T{
An experiment was also conducted with word frequencies via \textit{Google Ngram}\footnote{\url{https://books.google.com/ngrams}} (\M{see} Section \ref{subsec:wordfrequ} \M{in the appendix}).}

\subsection{Model-related Observations}

The basic expectation that the cosine measure would generally perform the worst and the FFN classifier the best was met (see Tables~\ref{tab:static-score-summary}--\ref{tab:gpt-score-summary}).
Interestingly, cosine is also outperformed by distance correlation in almost all cases.

Among the static models, GloVe and fastText performed best in most cases, especially on the room and part dataset (Table \ref{tab:static-score-summary}).
Although Levy performs by far the worst in the room dataset, it keeps up with all classification results in the verb dataset.
One reason for this could be the dependency-based learning strategy, which seems to work very well for verb associations, even though it was trained on a much smaller data set.

Among the masked-language models, BERT-Base surprisingly performed the best (Table~\ref{tab:masked-score-summary}).
BERT-Large achieved the better Increased Log Probabilities, but the FFN classifier still worked better with the vector representations of the Base variant.
This suggests that although associations are represented in a more fine-grained manner in BERT-Large, they are more difficult to retrieve due to the size of this model.

Among the masked-language models, GPT-J \T{(which was trained \M{with by far the largest} training data)} performs best (Table~\ref{tab:gpt-score-summary}).
Context-based models generally seem to determine the target given the source ($P(\textit{target}\mid \textit{source})$) more easily than the reverse ($P(\textit{source}\mid \textit{target})$).
With verbs, on the other hand, the reverse effect occurs.
The GPT models show that the results for sources are better when weighted, while for targets the results are better without weighting.

In general, the SVM performed surprisingly well, even though only a linear kernel was used. 
But also the KNN method mostly performed better than the similarity measures.
However, FFN performs best in all cases, outperforming cosine (worst case) by increases in the interval $[6\%, 52\%]$ and outperforming the KNN approach (worst classifier) in each case by increases in the interval $[2\%, 43\%]$.
%


\subsection{Dataset-related Observations}

In terms of rooms, \textit{bathroom} scores the best, while \textit{living room} or \textit{office} usually score the worst.
This may be because many bathroom objects are related to specific bathroom activities (e.g., toothbrush, bathtub), while objects that used to be located in the living room are increasingly found in other rooms (e.g., television in the bedroom).
This would also explain why the results for \textit{kitchen} are also better.

On the part dataset, the static models actually performed significantly better than the contextualized models. 
This relates especially to GloVe and fastText which outperformed almost all contextualized models. 
%
%
Thus, static models are in some cases a good alternative to their contextualized counterparts.
However, the more technical the objects become (here \textit{mortise lock} and \textit{dishwasher}), the worse the results become.

On the verb dataset, the contextualized models perform minimally better.
As mentioned earlier, these models can associate objects with verbs more easily than the other way around.
Here, the largest difference in performance is observed in the case of Levy, where the results are almost equal to those of the other models, probably due to the learning strategy based on dependency trees.

In summary, knowledge extraction using language models, whether static or contextualized, is more effective using classifiers than using similarity measures commonly used in the field of bias research:
there is potential for this type of knowledge extraction, but at the price of training classifiers -- if one uses similarity measures instead, this knowledge is mostly out of reach.

\subsection{Relation Observation}

    \begin{figure}
        \centering
        \begin{subfigure}[b]{0.4\textwidth}
            \centering
            \includegraphics[width=\textwidth, trim={0 0 0 1.2cm},clip]{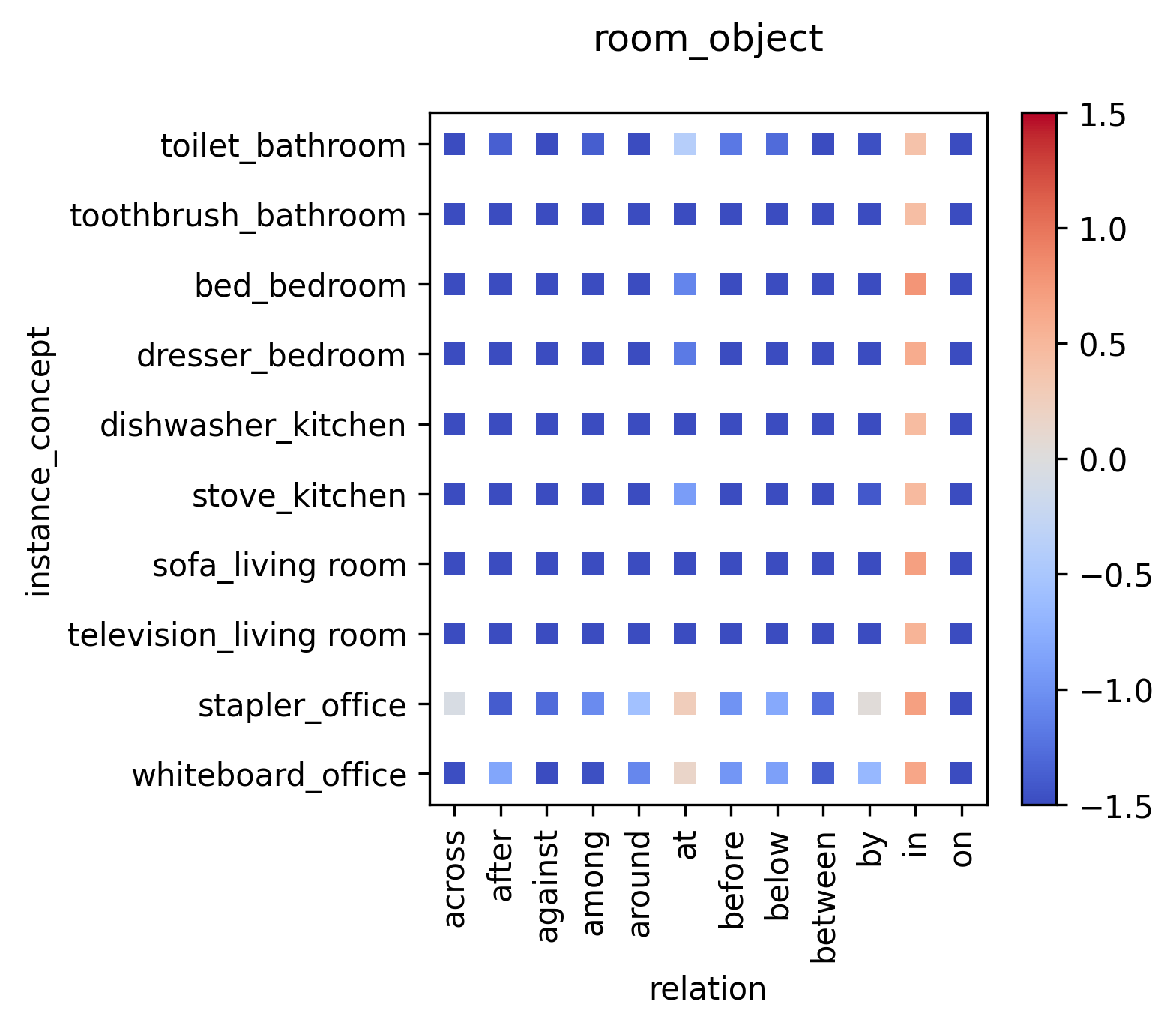}
        \end{subfigure}
        \begin{subfigure}[b]{0.4\textwidth}  
            \centering 
            \includegraphics[width=\textwidth, trim={0 0 0 1.2cm},clip]{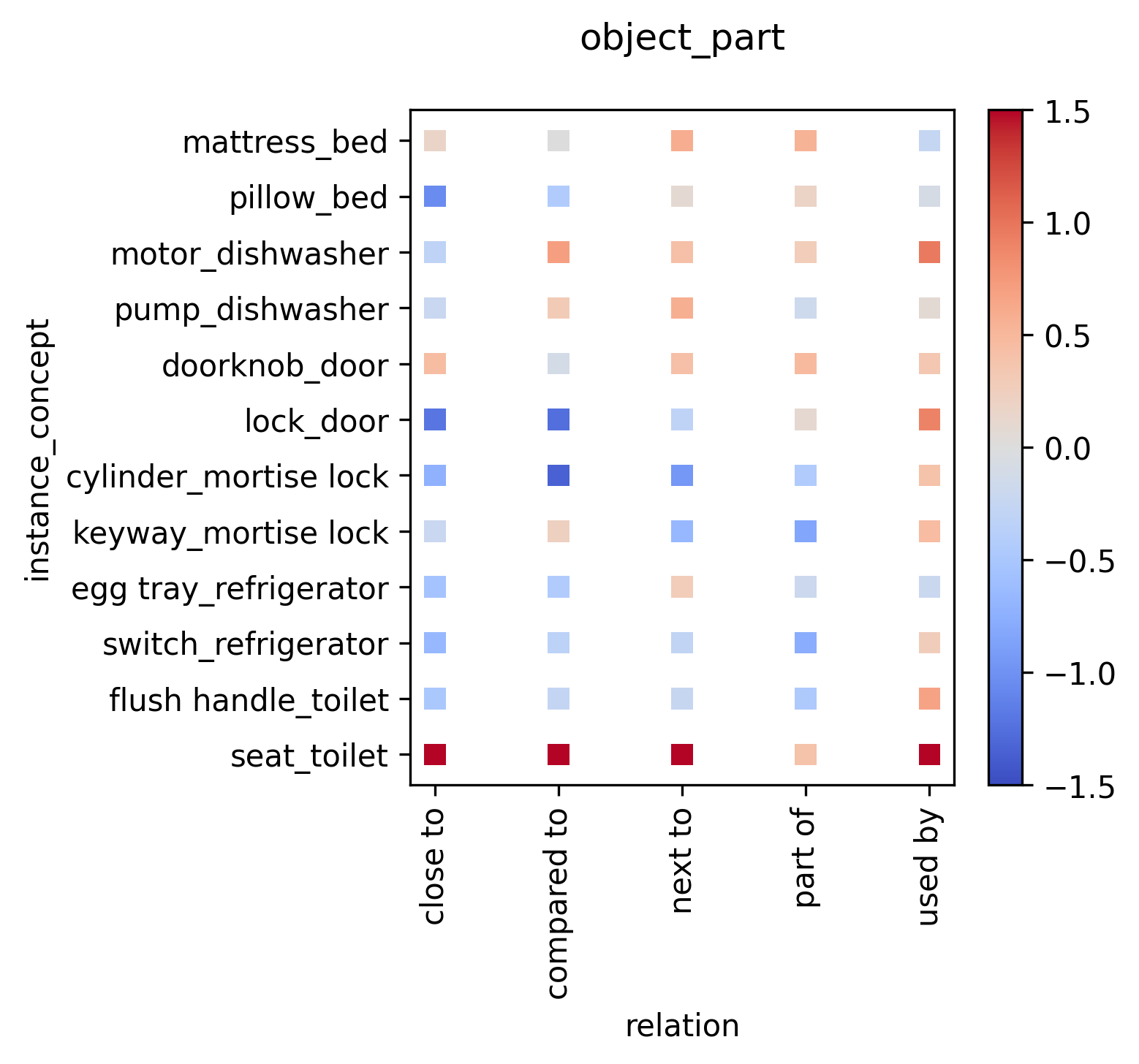}
        \end{subfigure}
    \caption{\T{Small relation evaluation of BERT-large after the method of~\newcite{Kurita:et:al:2019}.}}
     \label{fig:relations}
    \end{figure}

\T{\M{All previous evaluations only examined associations between instances and concepts, but not whether the models represent their true relations.}
To \M{fill this gap}, we repeated the experimental setup of ~\newcite{Kurita:et:al:2019} for the room and part dataset on BERT-large, but this time masked the \M{relation}.
The results are shown in Figure \ref{fig:relations}.
%
\M{Our selection of relations does not claim to be exhaustive, but serves as an illustration.}
\M{It shows that} while BERT-large is still very good at assigning objects \textit{in} rooms, the dominant relation \M{predicted} for parts is \textit{used by}. 
%
This suggests that BERT has problems correctly assigning object parts, an observation that could explain its poorer results while being consistent with findings of \cite{Lin:et:al:2020} (e.g., regarding counting parts).}

\section{Discussion} \label{sec:Discussion}
As good as the results obtained using classifiers are, they must be viewed with caution.
One can attribute their success to the fine-tuning of numerous parameters (and ultimately to overfitting); however, one can also attribute this success to nonlinear structuring of the information encoded in language models.
In other words, these models appear to encode object knowledge, but require a sophisticated apparatus to retrieve it.
\M{Thus, they should not be considered as an alternative to unsupervised approaches.}

Another issue is that our experiments do not yet allow for a comparison of model \textit{architectures}, as the models studied differ significantly in terms of the size of their parameter spaces and training data. 
Our experiments do suggest that certain smaller models come close to or even outperform the results of larger models. 
However, a comparison of model architectures would require controlling for these parameters.
Nevertheless, the results we have obtained are, in part, promising enough to encourage such research.

Finally, our experiments show that static models can perform better than contextualized models to some extent.
This finding is conditioned by our experiments and their context of application. 
\T{These observations that \textit{older} models perform better on certain tasks are consistent with other work (e.g. LSTMs on small datasets for intent classification~\cite{Ezen-Can:2020} or definiteness prediction~\cite{Kabbara:Cheung:2021}.}
At this point, a much broader analysis is needed (considering more areas and object relations), which also exploits contextual knowledge represented in contextualized models more than has been done here and in related work.
Nevertheless, it is generally difficult to obtain data for such a broader analysis, and our experiments are already broader in scope and consider finer relationships than similar approaches.

\section{Conclusion} \label{sec:Conclusion_Futurework}

We evaluated static and contextualized models as potential resources for object-related knowledge extraction.
To this end, we examined three datasets (to identify typical artifacts in rooms, objects of actions, or parts of objects).
We also experimented with different similarity measures and classifiers to extract the information contained in the language models.
In doing so, we have shown that the models in combination with the measures differ greatly in terms of the amount of knowledge they allow \M{for} \T{extracting}.
There is a weak trend that BERT-Base is the best performer among contextualized models, and GloVe and fastText among static models.
Secondly, our results suggest that approaches based on classifiers perform significantly better than similarity measures.
\T{Thirdly, we have shown that static models perform almost as well as contextualized models -- in some cases even better.} 
%
%
This result shows that research on these models needs to be advanced. 
\T{In future \M{work} we will also investigate how grounded language models perform on such datasets.}
%
%
However, as noted above, this requires a significant expansion of bias research, such as that conducted here to enable knowledge extraction.

\section*{Acknowledgements}
The support by the \textit{Stiftung Polytechnische Gesellschaft} (SPTG) is gratefully acknowledged.

\bibliography{bib}
\bibliographystyle{template_acl/acl_natbib}
\appendix

\section{Appendix}
\begin{sidewaysfigure*}
\begin{subfigure}{0.25\hsize}\centering
    \includegraphics[width=0.9\hsize, trim={0 0.6cm 0 1.2cm},clip]{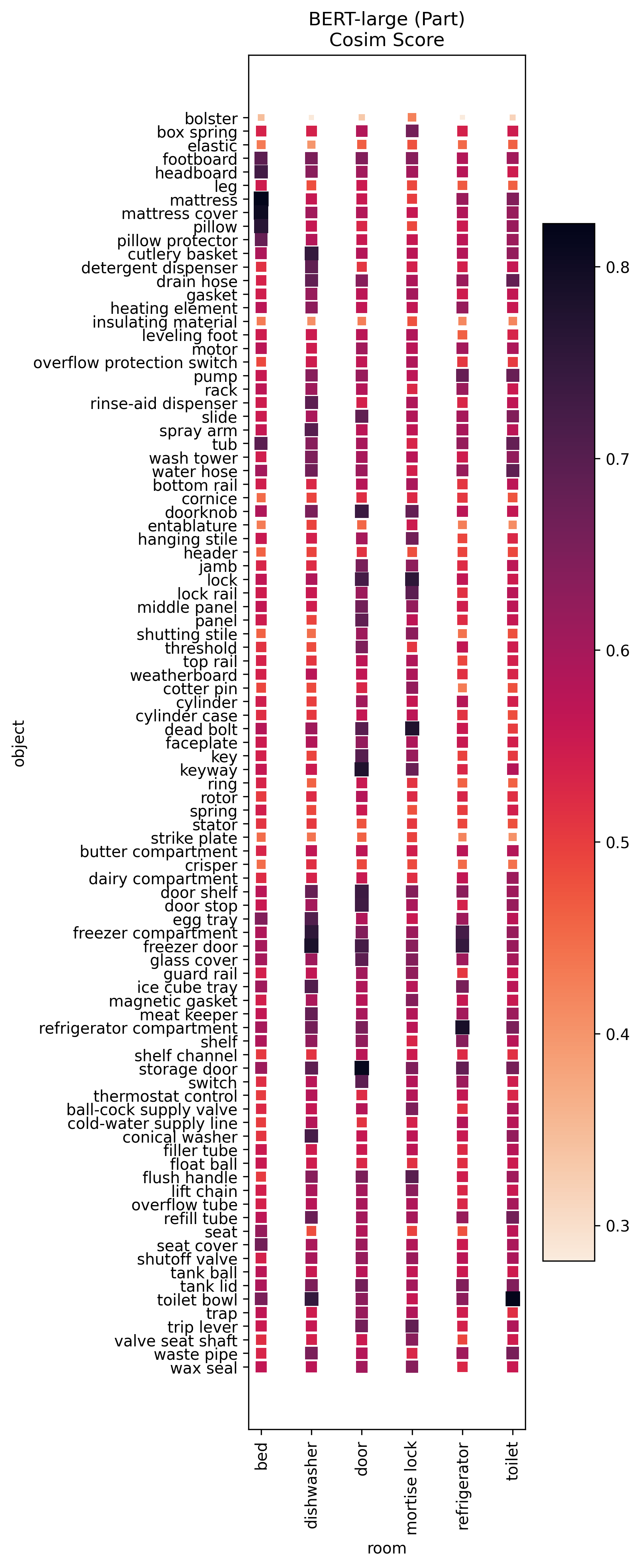}
    \caption{Cosine Score}
\end{subfigure}%
\begin{subfigure}{0.25\hsize}\centering
    \includegraphics[width=0.9\hsize, trim={0 0.6cm 0 1.2cm},clip]{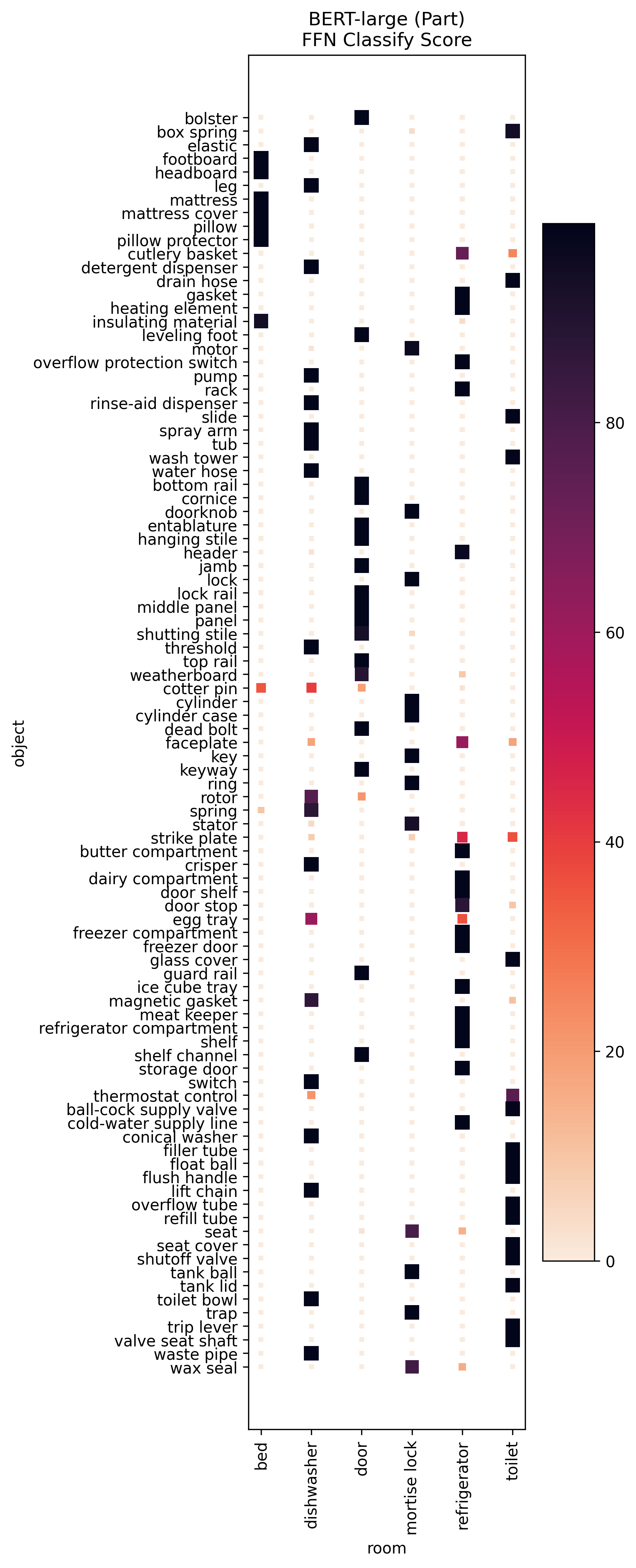}
    \caption{FFN Classify Score}
\end{subfigure}
\begin{subfigure}{0.25\hsize}\centering
    \includegraphics[width=0.9\hsize, trim={0 0.6cm 0 1.2cm},clip]{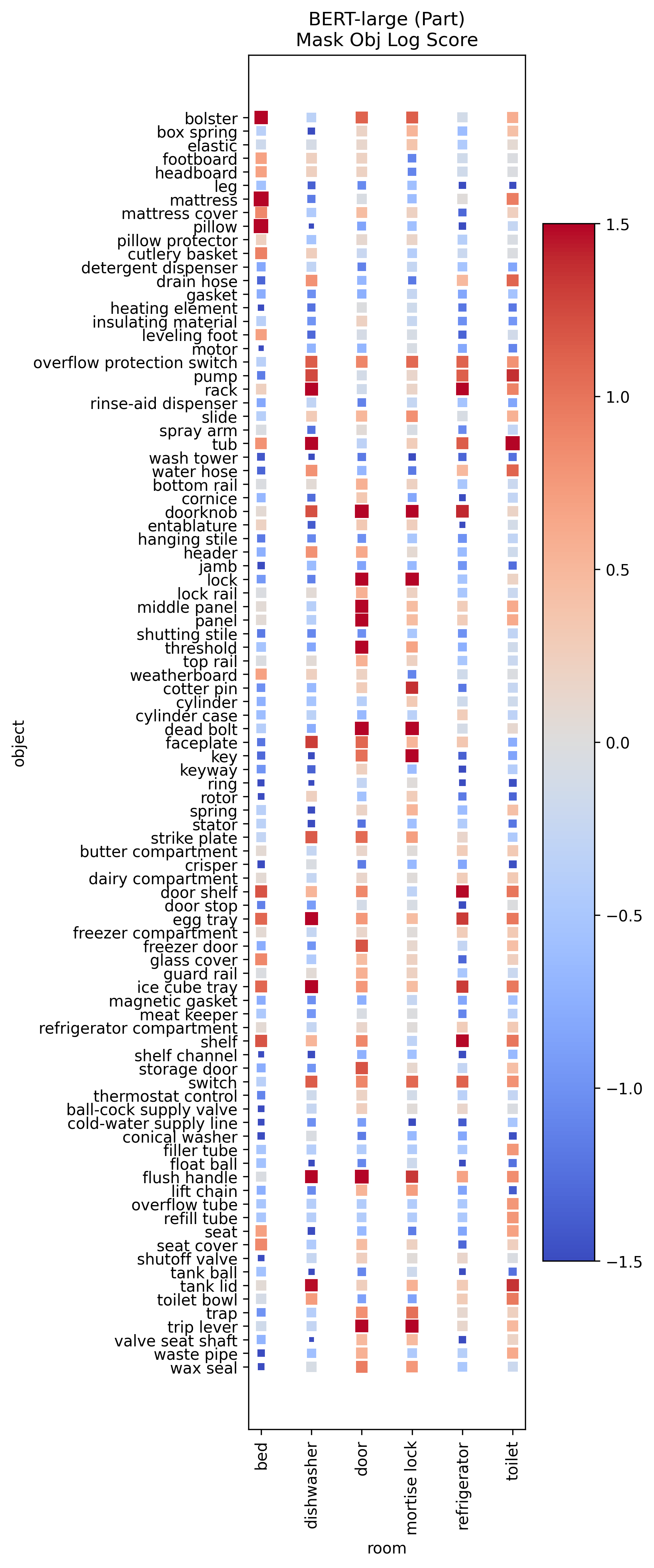}
    \caption{Mask Part Score}
\end{subfigure}
\begin{subfigure}{0.25\hsize}\centering
    \includegraphics[width=0.9\hsize, trim={0 0.6cm 0 1.2cm},clip]{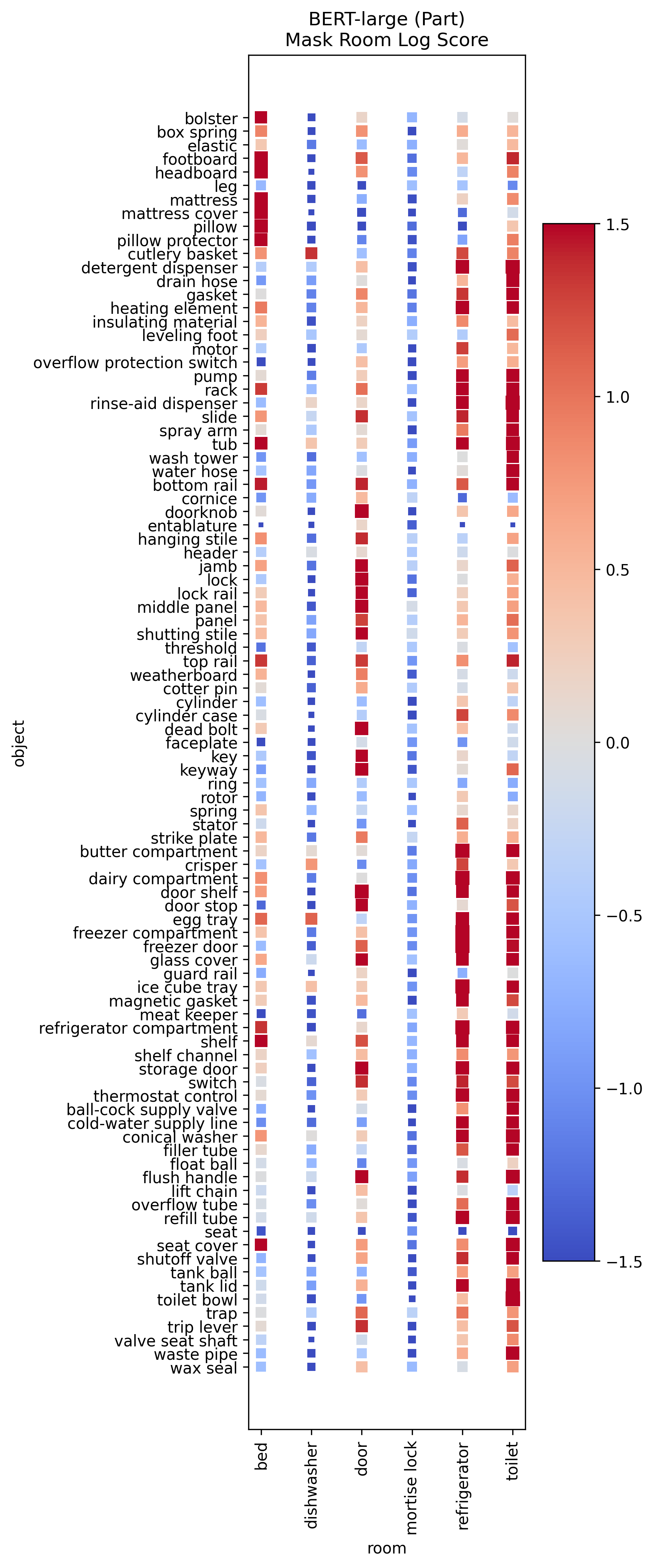}
    \caption{Mask Object Score}
\end{subfigure}
\caption{Association heatmap of BERT-Large on the part dataset. The parts (sources) on the y-axis are grouped by the room in which they are most likely to be located according to the \textit{Online-Bildwörterbuch Dataset}.}
\label{fig:bert-part-all}
\end{sidewaysfigure*}

\begin{figure*}
    \centering
    \begin{subfigure}[b]{0.4\textwidth}
        \centering
        \includegraphics[width=\textwidth, trim={0 0.6cm 0 1.2cm},clip]{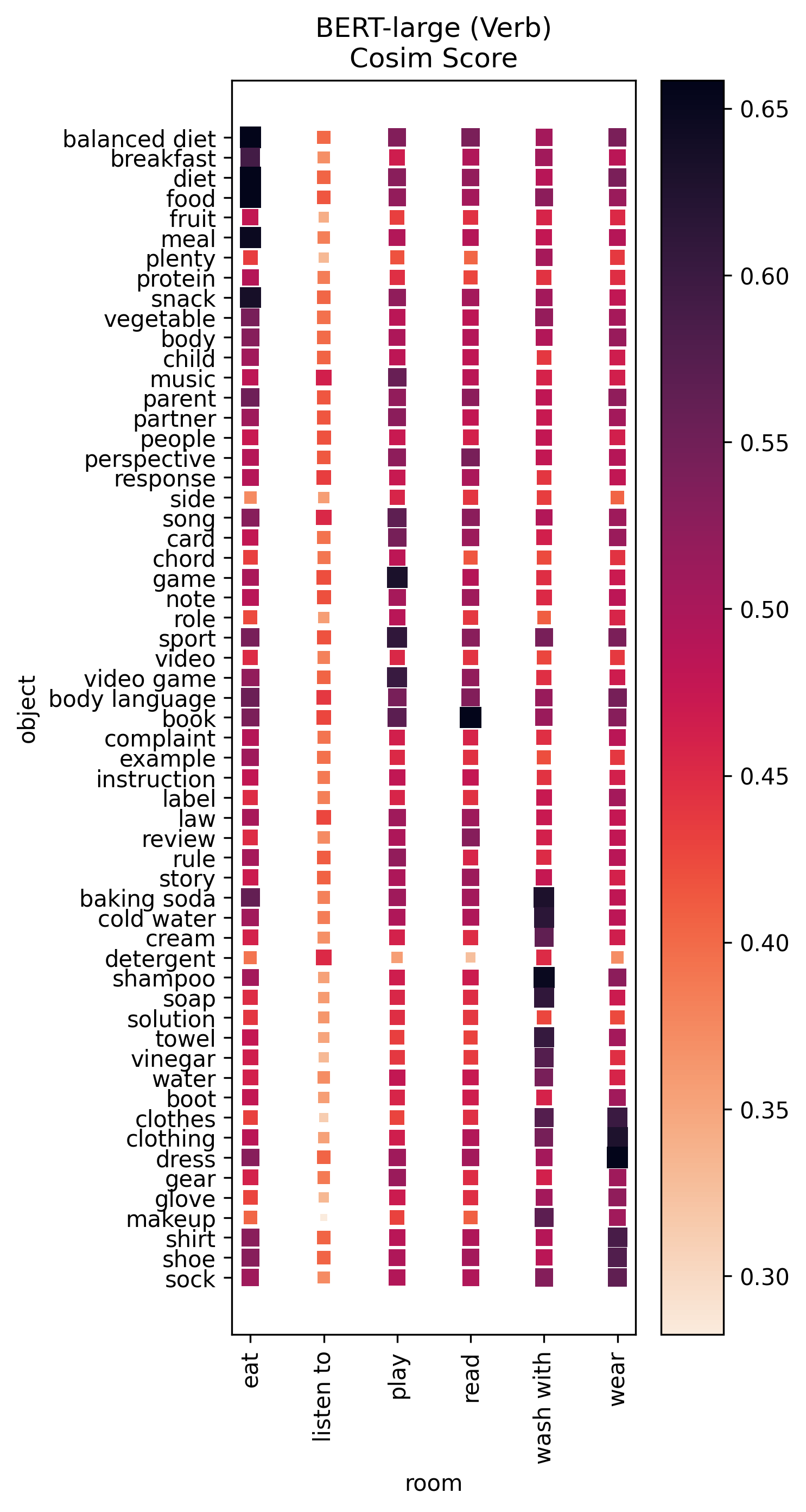}
        \vspace{-0.8cm}
        \caption{Cosine Score}
    \end{subfigure}
    \hfill
    \begin{subfigure}[b]{0.4\textwidth}  
        \centering 
        \includegraphics[width=\textwidth, trim={0 0.6cm 0 1.2cm},clip]{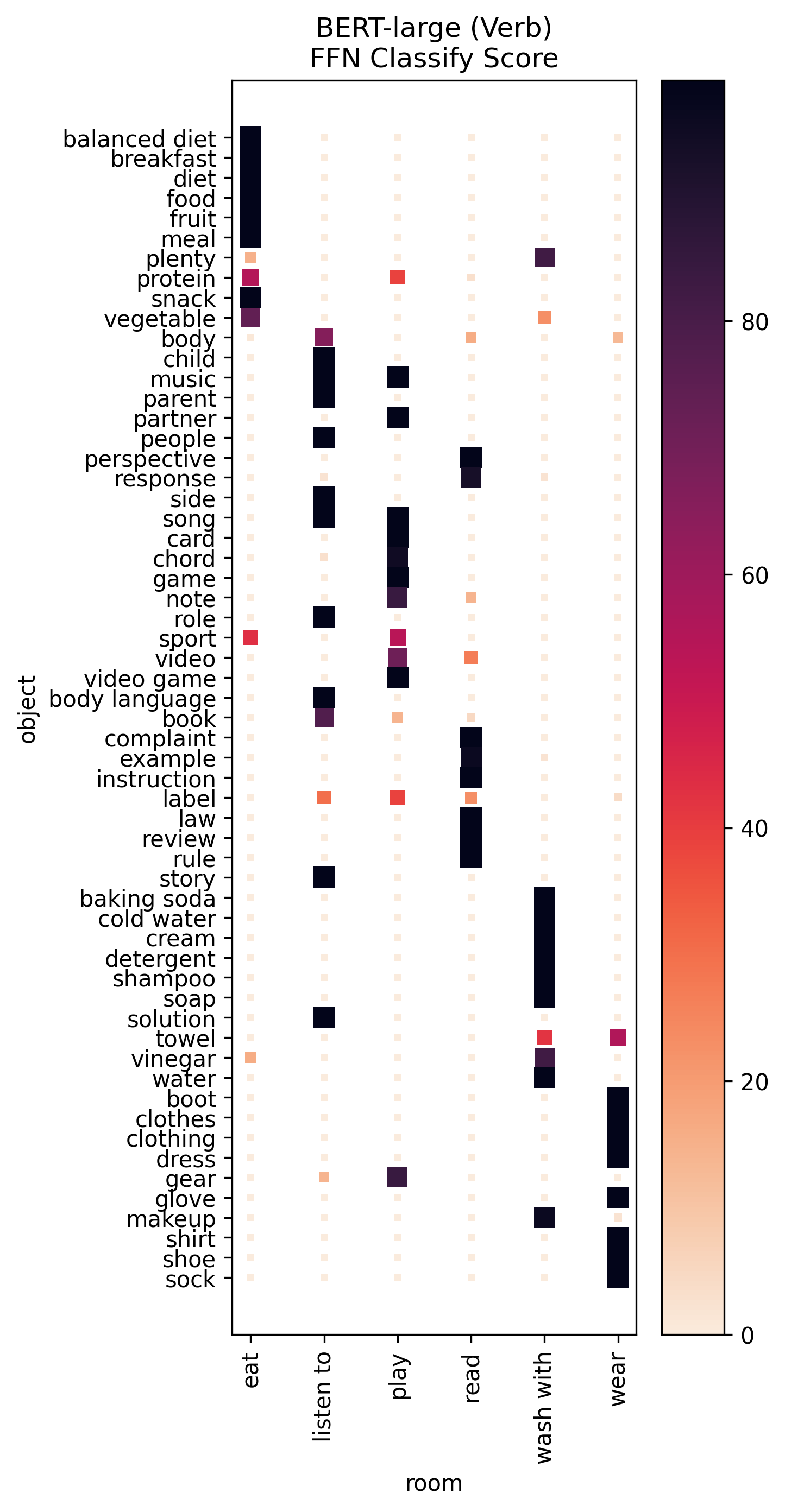}
        \vspace{-0.8cm}
        \caption{FFN Classify Score}
    \end{subfigure}
    \vskip\baselineskip
    \begin{subfigure}[b]{0.4\textwidth}   
        \centering 
        \includegraphics[width=\textwidth, trim={0 0.6cm 0 1.2cm},clip]{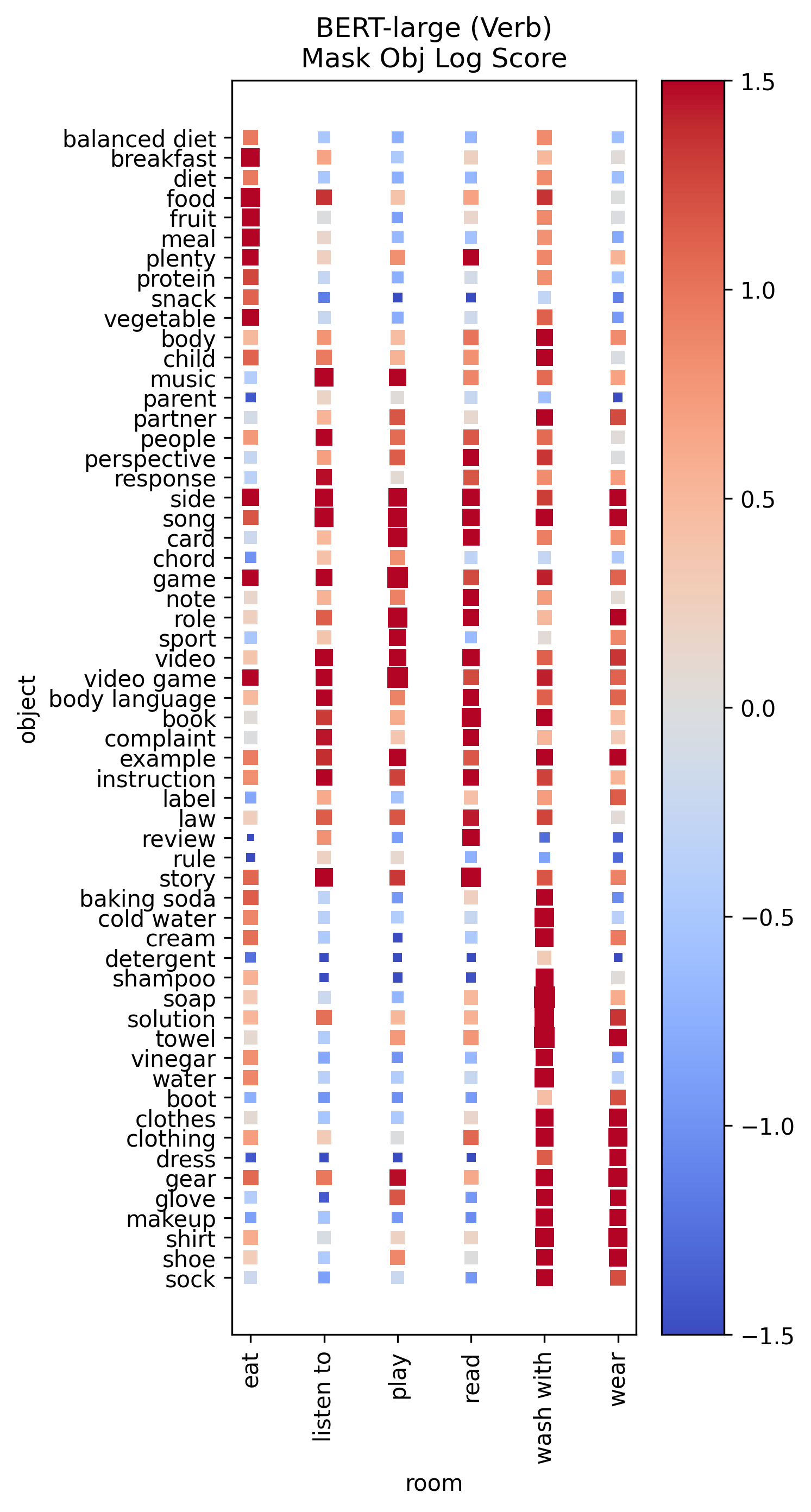}
        \vspace{-0.8cm}
        \caption{Mask Object Score}
    \end{subfigure}
    \hfill
    \begin{subfigure}[b]{0.4\textwidth}   
        \centering 
        \includegraphics[width=\textwidth, trim={0 0.6cm 0 1.2cm},clip]{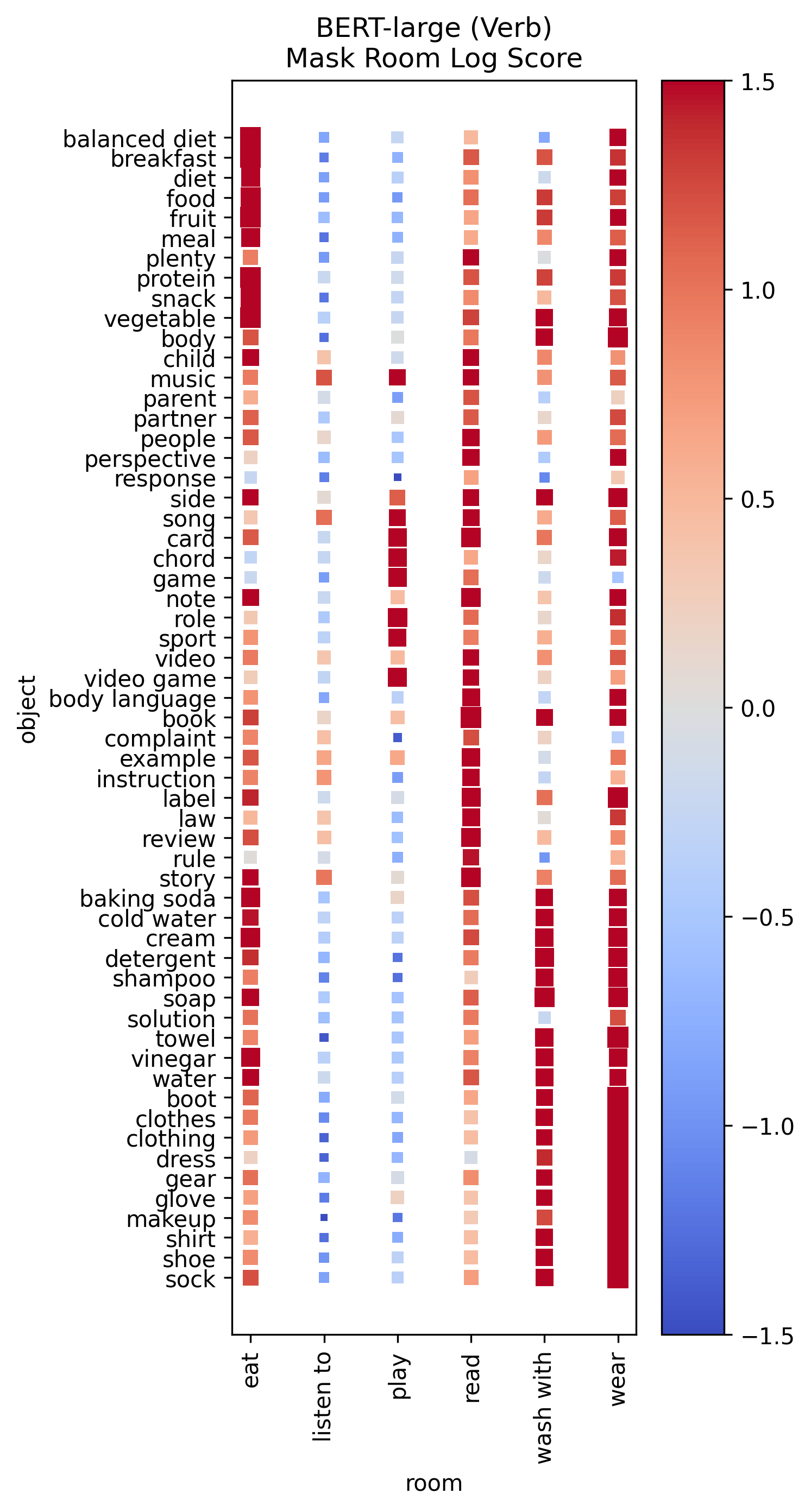}
        \vspace{-0.8cm}
        \caption{Mask Verb Score}
    \end{subfigure}
    \caption{Association heatmap of BERT-Large on the verb dataset. The objects (sources) on the y-axis are grouped by the room in which they are most likely to be located according to the \textit{HowToKB Dataset}.}
    \label{fig:bert-verb-all}
\end{figure*}

A tabular breakdown of the datasets used can be seen in Table \ref{tab:room-obj}, \ref{tab:obj-part} and \ref{tab:verb-obj}.
The exact models used are listed in Table~\ref{tab:model-overview}.
The heatmap visualizations for the other two datasets are in Figure \ref{fig:bert-part-all} and \ref{fig:bert-verb-all}.

\subsection{Word Frequency} \label{subsec:wordfrequ}
\T{We also conducted an experiment to correlate the scores with their frequency.
For this purpose, the corresponding objects of each target were selected.
And then the distance correlation between the scores and the corresponding word frequency was calculated based on the average of the last 10 years of \textit{Google Ngrams}.
The results are shown in Table \ref{tab:room_obj_g-ngram}.
The correlations are not particularly significant (mostly p $\geq$ 0.1), but it is noticeable that especially the cosine score depends strongly on the word frequency.
The classifiers are generally less sensitive.
}

\begin{table*}[!hpt]
    \setlength{\tabcolsep}{3pt}
    \centering
    \scriptsize
    \begin{tabularx}{\linewidth}{ X  c | X  c | X  c | X  c | X  c }
    \toprule
    \multicolumn{2}{ c |}{\textbf{bathroom}} & \multicolumn{2}{ c |}{\textbf{bedroom}} & \multicolumn{2}{ c |}{\textbf{kitchen}} & \multicolumn{2}{ c |}{\textbf{living room}} & \multicolumn{2}{ c }{\textbf{office}} \\ 
    object & score & object & score & object & score & object & score & object & score \\ \midrule
    
    toilet & 1.00 & dresser & 1.00 & drying rack & 1.00 & coffee table & 0.94 & whiteboard & 1.00 \\
    bathtub & 1.00 & night stand & 1.00 & kitchen island & 1.00 & ottoman & 0.93 & room divider & 0.94 \\
    toothbrush holder & 1.00 & headboard & 1.00 & pot & 1.00 & fireplace & 0.87 & stapler & 0.92 \\
    toothpaste & 1.00 & bed & 0.97 & frying pan & 1.00 & dvd player & 0.69 & cork board & 0.92 \\
    shower curtain & 1.00 & alarm clock & 0.97 & spice rack & 1.00 & sofa & 0.68 & file & 0.88 \\
    toothbrush & 0.97 & laundry basket & 0.86 & cutting board & 1.00 & decorative plate & 0.61 & keyboard & 0.85 \\
    towel rod & 0.96 & hat & 0.74 & blender & 1.00 & tv stand & 0.57 & mouse & 0.84 \\
    toilet paper & 0.96 & doll & 0.70 & knife & 1.00 & blanket & 0.55 & pen & 0.83 \\
    squeeze tube & 0.95 & stuffed animal & 0.60 & stove & 0.98 & television & 0.53 & computer & 0.82 \\
    faucet handle & 0.82 & pillow & 0.56 & dishwasher & 0.97 & remote control & 0.50 & column & 0.81 \\ \bottomrule

    \end{tabularx}
    \caption{Statistics generated from ScanNet using NYU categories: 
    %
    \textit{score} is the conditional probability $P(\mathit{room} \mid \mathit{object})$ of the room given the object based on the frequencies observable in NYU.}
    \label{tab:room-obj}
\end{table*}
\newcolumntype{M}{>{\begin{varwidth}{1.5cm}}l<{\end{varwidth}}}

\begin{table*}[!hpt]
    \centering
     \setlength{\tabcolsep}{3pt}
    \tiny
    \begin{tabularx}{\linewidth}{ X  c | X  c | X  c | X  c | X  c | X  c }
    \toprule
    \multicolumn{2}{ c |}{\textbf{bed}} & \multicolumn{2}{ c |}{\textbf{dishwasher}} & \multicolumn{2}{ c |}{\textbf{door}} & \multicolumn{2}{ c |}{\textbf{mortise lock}} & \multicolumn{2}{ c }{\textbf{refrigerator}} & \multicolumn{2}{ c }{\textbf{toilet}}\\ 
    object & score & object & score & object & score & object & score & object & score & object & score \\ \midrule
    
    pillow & 1.00 & drain hose & 1.00 & lock & 1.00 & ring & 1.00 & switch & 1.00 & valve seat shaft & 1.00 \\ 
    bolster & 1.00 & overflow protection switch & 1.00 & cornice & 1.00 & keyway & 1.00 & refrigerator compartment & 1.00 & tank lid & 1.00 \\
    mattress cover & 1.00 & tub & 1.00 & hanging stile & 1.00 & cotter pin & 1.00 & egg tray & 1.00 & conical washer & 1.00 \\
    leg & 1.00 & pump & 1.00 & entablature & 1.00 & spring & 1.00 & shelf channel & 1.00 & lift chain & 1.00 \\
    box spring & 1.00 & gasket & 1.00 & top rail & 1.00 & rotor & 1.00 & magnetic gasket & 1.00 & seat & 1.00 \\
    headboard & 1.00 & water hose & 1.00 & middle panel & 1.00 & cylinder case & 1.00 & storage door & 1.00 & shutoff valve & 1.00 \\
    mattress & 1.00 & heating element & 1.00 & bottom rail & 1.00 & key & 1.00 & freezer door & 1.00 & trip lever & 1.00 \\
    pillow protector & 1.00 & rack & 1.00 & panel & 1.00 & faceplate & 1.00 & guard rail & 1.00 & ball-cock supply valve & 1.00 \\
    elastic & 1.00 & cutlery basket & 1.00 & jamb & 1.00 & dead bolt & 1.00 & crisper & 1.00 & toilet bowl & 1.00 \\
    footboard & 1.00 & wash tower & 1.00 & doorknob & 1.00 & cylinder & 1.00 & glass cover & 1.00 & flush handle & 1.00 \\
  &   & motor & 1.00 & threshold & 1.00 & stator & 1.00 & butter compartment & 1.00 & wax seal & 1.00 \\
  &   & detergent dispenser & 1.00 & weatherboard & 1.00 & strike plate & 1.00 & thermostat control & 1.00 & tank ball & 1.00 \\
  &   & slide & 1.00 & lock rail & 1.00 &   &   & freezer compartment & 1.00 & float ball & 1.00 \\
  &   & leveling foot & 1.00 & shutting stile & 1.00 &   &   & ice cube tray & 1.00 & filler tube & 1.00 \\
  &   & insulating material & 1.00 & header & 1.00 &   &   & meat keeper & 1.00 & waste pipe & 1.00 \\
  &   & spray arm & 1.00 &   &   &   &   & door stop & 1.00 & seat cover & 1.00 \\
  &   & rinse-aid dispenser & 1.00 &   &   &   &   & shelf & 1.00 & cold-water supply line & 1.00 \\
  &   &   &   &   &   &   &   & dairy compartment & 1.00 & overflow tube & 1.00 \\
  &   &   &   &   &   &   &   & door shelf & 1.00 & trap & 1.00 \\
  &   &   &   &   &   &   &   &   &   & refill tube & 1.00 \\ \bottomrule

    \end{tabularx}
    \caption{A subset of part-whole relations extracted from \textit{Online-Bildwörterbuch}. \T{All parts have a value of \textit{1.00} in our data set, because they only occur with this object.}}
    \label{tab:obj-part}
\end{table*}
\begin{table*}[!hpt]
    \centering
     \setlength{\tabcolsep}{3pt}
    \scriptsize
    \begin{tabularx}{\linewidth}{ X  c | X  c | X  c | X  c | X  c | X  c }
    \toprule
    \multicolumn{2}{ c |}{\textbf{eat}} & \multicolumn{2}{ c |}{\textbf{listen to}} & \multicolumn{2}{ c |}{\textbf{play}} & \multicolumn{2}{ c |}{\textbf{read}} & \multicolumn{2}{ c }{\textbf{wash with}} & \multicolumn{2}{ c }{\textbf{wear}}\\ 
    object & score & object & score & object & score & object & score & object & score & object & score \\ \midrule
    
    food & 0.13 & music & 0.22 & game & 0.27 & book & 0.08 & soap & 0.29 & clothing & 0.07 \\
    diet & 0.08 & song & 0.03 & music & 0.06 & label & 0.06 & water & 0.29 & glove & 0.06 \\
    meal & 0.07 & body & 0.03 & note & 0.04 & instruction & 0.05 & vinegar & 0.04 & shoe & 0.05 \\
    breakfast & 0.04 & side & 0.02 & sport & 0.03 & review & 0.04 & solution & 0.03 & clothes & 0.05 \\
    balanced diet & 0.03 & partner & 0.02 & chord & 0.02 & body language & 0.02 & detergent & 0.03 & shirt & 0.02 \\
    fruit & 0.03 & child & 0.02 & song & 0.02 & rule & 0.02 & baking soda & 0.03 & makeup & 0.02 \\
    vegetable & 0.03 & perspective & 0.02 & video game & 0.02 & example & 0.02 & cream & 0.02 & gear & 0.02 \\
    plenty & 0.03 & response & 0.02 & card & 0.02 & complaint & 0.01 & shampoo & 0.02 & boot & 0.02 \\
    protein & 0.03 & parent & 0.02 & role & 0.02 & law & 0.01 & towel & 0.02 & dress & 0.02 \\
    snack & 0.02 & people & 0.02 & video & 0.02 & story & 0.01 & cold water & 0.02 & sock & 0.02 \\ \bottomrule
    \end{tabularx}
    \caption{A subset of verb-object relations extracted from an updated version of HowToKB.}
    \label{tab:verb-obj}
\end{table*}

\newcolumntype{P}[1]{>{\centering\arraybackslash}p{#1}}
\begin{table*}[!hpt]
    \centering
     \setlength{\tabcolsep}{3pt}
    \scriptsize
    \begin{tabularx}{\linewidth}{l | l | c | c | P{1.4cm} | X}
    \toprule
    Model & Specification & Dimension & Parameters & Dataset Size (T ; S) & URL \\ \midrule
    word2vec & GoogleNews-vectors-negative300 & 300 & - & 100B ; - & \url{https://code.google.com/archive/p/word2vec/} \\
    Glove & Common Crawl - glove.840B.300d & 300 & - & 840B ; - & \url{https://nlp.stanford.edu/projects/glove/} \\
    Levy & Dependency-Based Words & 300 & - & English Wiki ($\sim$ 2B tokens) & \url{https://levyomer.wordpress.com/2014/04/25/dependency-based-word-embeddings/} \\
    fastText & crawl-300d-2M-subword & 300 & - & 600B ; - & \url{https://fasttext.cc/docs/en/english-vectors.html} \\
    static-BERT & bert\_12layer\_sent & 768 & - & +1.28B ; - & \url{https://zenodo.org/record/5055755} \\
    \midrule
    
    BERT-Base & bert-base-uncased & 768 & $\sim$ 110M & 3.3B ; 16GB & \url{https://huggingface.co/bert-base-uncased} \\
    BERT-Large & bert-large-uncased & 1024 & $\sim$ 336M & 3.3B ; 16GB & \url{https://huggingface.co/bert-large-uncased} \\
    RoBERTa & roberta-large & 1024 & $\sim$ 336M & - ; 160GB & \url{https://huggingface.co/roberta-large} \\
    ELECTRA & electra-large-generator & 256 & $\sim$ 51M & & \url{https://huggingface.co/google/electra-large-generator} \\
    ALBERT & albert-xxlarge-v2 & 4096 & $\sim$ 223M & 3.3B ; 16GB & \url{https://huggingface.co/albert-xxlarge-v2} \\\midrule
    
    GPT2 & gpt2-large & 1280 & $\sim$ 774M & - ; 40GB & \url{https://huggingface.co/gpt2-large} \\
    GPT-Neo & gpt-neo-2.7B & 2560 & $\sim$ 2.7B & 420B ; - & \url{https://huggingface.co/EleutherAI/gpt-neo-2.7B} \\
    GPT-J & gpt-j-6B & 4096 & $\sim$ 6B & - ; 825GB & \url{https://huggingface.co/EleutherAI/gpt-j-6B} \\
    \bottomrule
    \end{tabularx}
    \caption{\T{Model overview. Mostly only the token quantity (T) or the dataset size (S) was given.}}
    \label{tab:model-overview}
\end{table*}

\begin{table*}[!hpt]
    \centering
    \small
    \begin{tabularx}{\linewidth}{c c c | X}
    \toprule
    Task & Model & Data & Templates \\ \midrule
    \multirow{12}{*}{\shortstack[c]{Cosine Score \\ \& \\ Classification}} & \multirow{12}{*}{MLM \& CLM} & \multirow{6}{*}{Room \& Objects \& Parts} & This is a/an \{x\}.\\
    & & & That is a/an \{x\}. \\
    & & & There is a/an \{x\}. \\
    & & & Here is a/an \{x\}. \\
    & & & A/An \{x\} is here. \\
    & & & A/An \{x\} is there. \\ \cmidrule{3-4}
    
    & & \multirow{6}{*}{Verbs} & I \{x\} something. \\
    & & & I \{x\} anything.\\
    & & & I \{x\}.\\
    & & & You \{x\} something.\\
    & & & You \{x\} anything.\\
    & & & You \{x\}.\\ \midrule
    
    \multirow{8}{*}{\shortstack[c]{Increased \\ Log Probability}} & \multirow{3}{*}{MLM} & Room \& Object & A/An \{obj\} is usually in the \{room\}. \\
     &  & Object \& Part & A/An \{part\} is usually part of a/an \{obj\}. \\
     &  & Verb \& Object & I usually \{verb\} this \{obj\}. \\ \cmidrule{2-4}
     
     & \multirow{5}{*}{CLM} & \multirow{2}{*}{Room \& Object} & A/An \{obj\} is usually in the ... \\
     & & & In the \{room\} is usually a/an ... \\
     
    & & \multirow{2}{*}{Object \& Part} & A/An \{part\} is usually part of a ...\\
    & & & In the \{obj\} is usually a/an ... \\
    
    & & Verb \& Object & I usually \{verb\} this ...\\
    \bottomrule
    \end{tabularx}
    \caption{Templates for calculating scores regarding \textit{Masked Language Models} (MLM) and \textit{Causal Language Models} (CLM). For more details, see Sec.~\ref{sec:Approach}.}
    \label{tab:context_template}
\end{table*}

\begin{table*}
    \setlength{\tabcolsep}{0.5pt}
    \centering
    \tiny
    \begin{tabularx}{\linewidth}{+l ^l | ^X ^X ^X ^X ^X ^X | ^X ^X ^X ^X ^X ^X | ^X ^X ^X ^X ^X ^X | ^X ^X ^X ^X ^X ^X | ^X ^X ^X ^X ^X ^X}
    \toprule
    &  & \multicolumn{6}{c}{\textbf{Word2Vec}} & \multicolumn{6}{c}{\textbf{GloVe}} & \multicolumn{6}{c}{\textbf{Levy}} & \multicolumn{6}{c}{\textbf{fastText}} & \multicolumn{6}{c}{\textbf{static-BERT}}\\ 
    &  & cos & dist & kend & knn & svm & ffn & cos & dist & kend & knn & svm & ffn & cos & dist & kend & knn & svm & ffn & cos & dist & kend & knn & svm & ffn & cos & dist & kend & knn & svm & ffn \\ \midrule
    \multirow{5}{*}{\textbf{\rotatebox[origin=c]{90}{Room}}} 
    & bathroom & 0.73* & 0.75* & 0.78* & 0.31 & 0.31 & 0.22 & 0.53 & 0.56* & 0.57* & 0.23 & 0.00 & 0.23 & 0.65* & 0.67* & 0.68* & 0.25 & 0.00 & 0.32 & 0.65* & 0.67* & 0.70* & 0.00 & 0.00 & 0.00 & 0.74* & 0.75* & 0.76* & 0.56* & 0.41 & 0.47 \\
    
    & bedroom & 0.55 & 0.53 & 0.56 & 0.00 & 0.36 & 0.35 & 0.72* & 0.72* & 0.69* & 0.50* & 0.35 & 0.57* & 0.51 & 0.51 & 0.50 & 0.00 & 0.00 & 0.37 & 0.58 & 0.54 & 0.53 & 0.66* & 0.55* & 0.35 & 0.45 & 0.44 & 0.44 & 0.68* & 0.35 & 0.36 \\
    
    & kitchen & 0.51 & 0.52 & 0.51 & 0.35 & 0.49 & 0.42 & 0.37 & 0.34 & 0.34 & 0.29 & 0.35 & 0.33 & 0.46 & 0.46 & 0.46 & 0.20 & 0.00 & 0.20 & 0.33 & 0.36 & 0.40 & 0.30 & 0.20 & 0.37 & 0.46 & 0.46 & 0.46 & 0.20 & 0.31 & 0.42 \\
    
    & living room & 0.63* & 0.61* & 0.61* & 0.36 & 0.46 & 0.50 & 0.48 & 0.62* & 0.63* & 0.44 & 0.29 & 0.30 & 0.53 & 0.57 & 0.57 & 0.41 & 0.41 & 0.54 & 0.39 & 0.57 & 0.59 & 0.28 & 0.29 & 0.27 & 0.52 & 0.54 & 0.53 & 0.00 & 0.30 & 0.52 \\
    
    & office & 0.65* & 0.65* & 0.58 & 0.38 & 0.26 & 0.37 & 0.52 & 0.57 & 0.56 & 0.38 & 0.43 & 0.47 & 0.66* & 0.66* & 0.65* & 0.27 & 0.50 & 0.50 & 0.45 & 0.43 & 0.41 & 0.58* & 0.35 & 0.30 & 0.37 & 0.40 & 0.40 & 0.57* & 0.17 & 0.42 \\ 
    
    \midrule

    &  & \multicolumn{6}{c}{\textbf{BERT-Base}} & \multicolumn{6}{c}{\textbf{BERT-Large}} & \multicolumn{6}{c}{\textbf{RoBERTa}} & \multicolumn{6}{c}{\textbf{ElectraGen}} & \multicolumn{6}{c}{\textbf{Albert}} \\ 

    &  & cos & m-s & m-t & knn & svm & ffn & cos & m-s & m-t & knn & svm & ffn & cos & m-s & m-t & knn & svm & ffn & cos & m-s & m-t & knn & svm & ffn & cos & m-s & m-t & knn & svm & ffn \\ \midrule
    
    \multirow{5}{*}{\textbf{\rotatebox[origin=c]{90}{Room}}} 
    & bathroom & 0.40 & 0.35 & 0.30 & 0.23 & 0.23 & 0.23 & 0.52 & 0.34 & 0.50 & 0.23 & 0.23 & 0.24 & 0.61 & 0.47 & 0.42 & 0.43 & 0.23 & 0.35 & 0.58 & 0.66* & 0.39 & 0.35 & 0.35 & 0.40 & 0.69* & 0.56 & 0.36 & 0.34 & 0.36 & 0.39 \\
    
    & bedroom & 0.61* & 0.47 & 0.41 & 0.45 & 0.28 & 0.37 & 0.63* & 0.36 & 0.37 & 0.42 & 0.19 & 0.36 & 0.67* & 0.56* & 0.33 & 0.42 & 0.28 & 0.41 & 0.41 & 0.53 & 0.62 & 0.68* & 0.55* & 0.50* & 0.54 & 0.58* & 0.36 & 0.18 & 0.31 & 0.47 \\
    
    & kitchen & 0.34 & 0.60 & 0.35 & 0.30 & 0.23 & 0.43 & 0.38 & 0.73* & 0.45 & 0.75* & 0.75* & 0.62* & 0.65* & 0.38 & 0.49 & 0.46 & 0.34 & 0.24 & 0.48 & 0.37 & 0.37 & 0.25 & 0.44 & 0.43 & 0.43 & 0.75* & 0.38 & 0.35 & 0.22 & 0.45 \\
    
    & living room & 0.52 & 0.57 & 0.56 & 0.25 & 0.20 & 0.32 & 0.47 & 0.47 & 0.54 & 0.36 & 0.36 & 0.39 & 0.43 & 0.51 & 0.45 & 0.51 & 0.27 & 0.27 & 0.41 & 0.36 & 0.45 & 0.44 & 0.49 & 0.47 & 0.38 & 0.54 & 0.47 & 0.19 & 0.30 & 0.34 \\
    
    & office & 0.68* & 0.56 & 0.64 & 0.35 & 0.44 & 0.56* & 0.64* & 0.75* & 0.67* & 0.26 & 0.58* & 0.45 & 0.48 & 0.49 & 0.53 & 0.27 & 0.28 & 0.47 & 0.50 & 0.70* & 0.55 & 0.59* & 0.55 & 0.25 & 0.72* & 0.54 & 0.61 & 0.58* & 0.37 & 0.36 \\ 
    
    \bottomrule
    \end{tabularx}
    \caption{\T{Distance Correlation calculated on the word frequencies of Google Ngram. (* indicates significant at p < 0.1)}}
    \label{tab:room_obj_g-ngram}
\end{table*}

    \begin{figure*}
        \centering
        \begin{subfigure}[b]{0.4\textwidth}
            \centering
            \includegraphics[width=\textwidth, trim={0 0 0 1.2cm},clip]{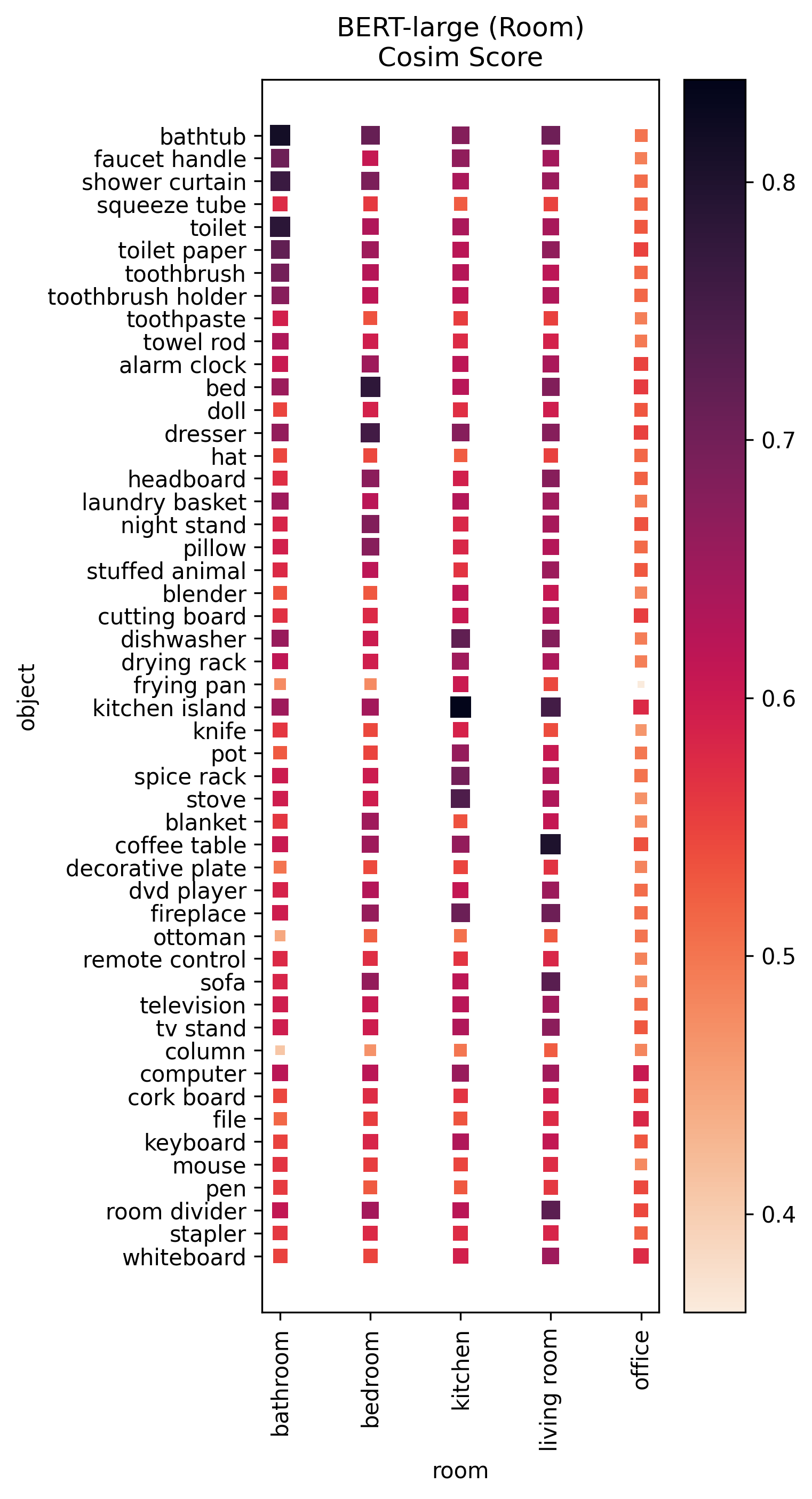}
            \vspace{-0.8cm}
            \caption{Cosine Score}
        \end{subfigure}
        \hfill
        \begin{subfigure}[b]{0.4\textwidth}  
            \centering 
            \includegraphics[width=\textwidth, trim={0 0 0 1.2cm},clip]{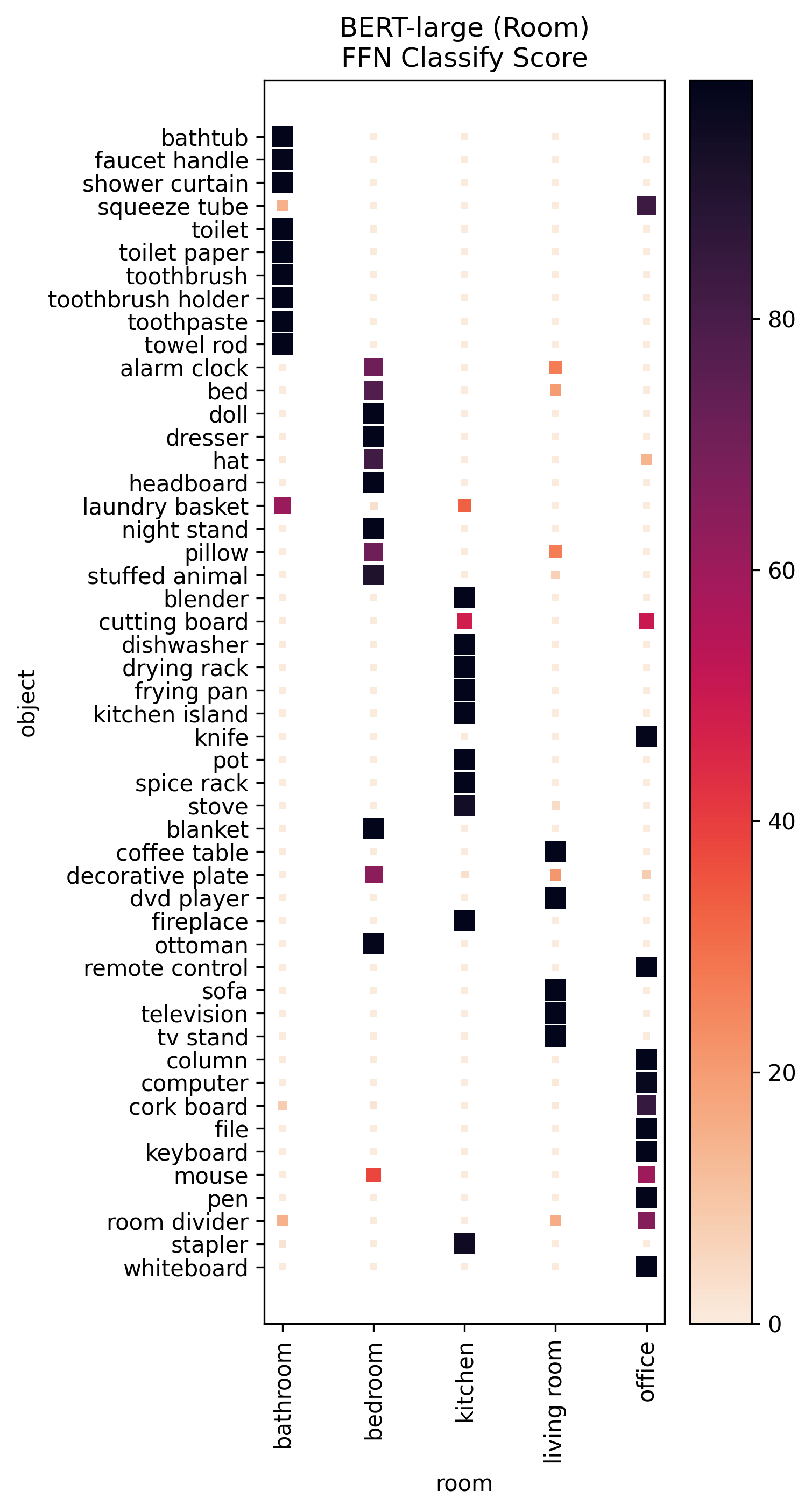}            \vspace{-0.8cm}
            \caption{FFN Classify Score}
        \end{subfigure}
        \vskip\baselineskip
        \begin{subfigure}[b]{0.4\textwidth}   
            \centering 
            \includegraphics[width=\textwidth, trim={0 0 0 1.2cm},clip]{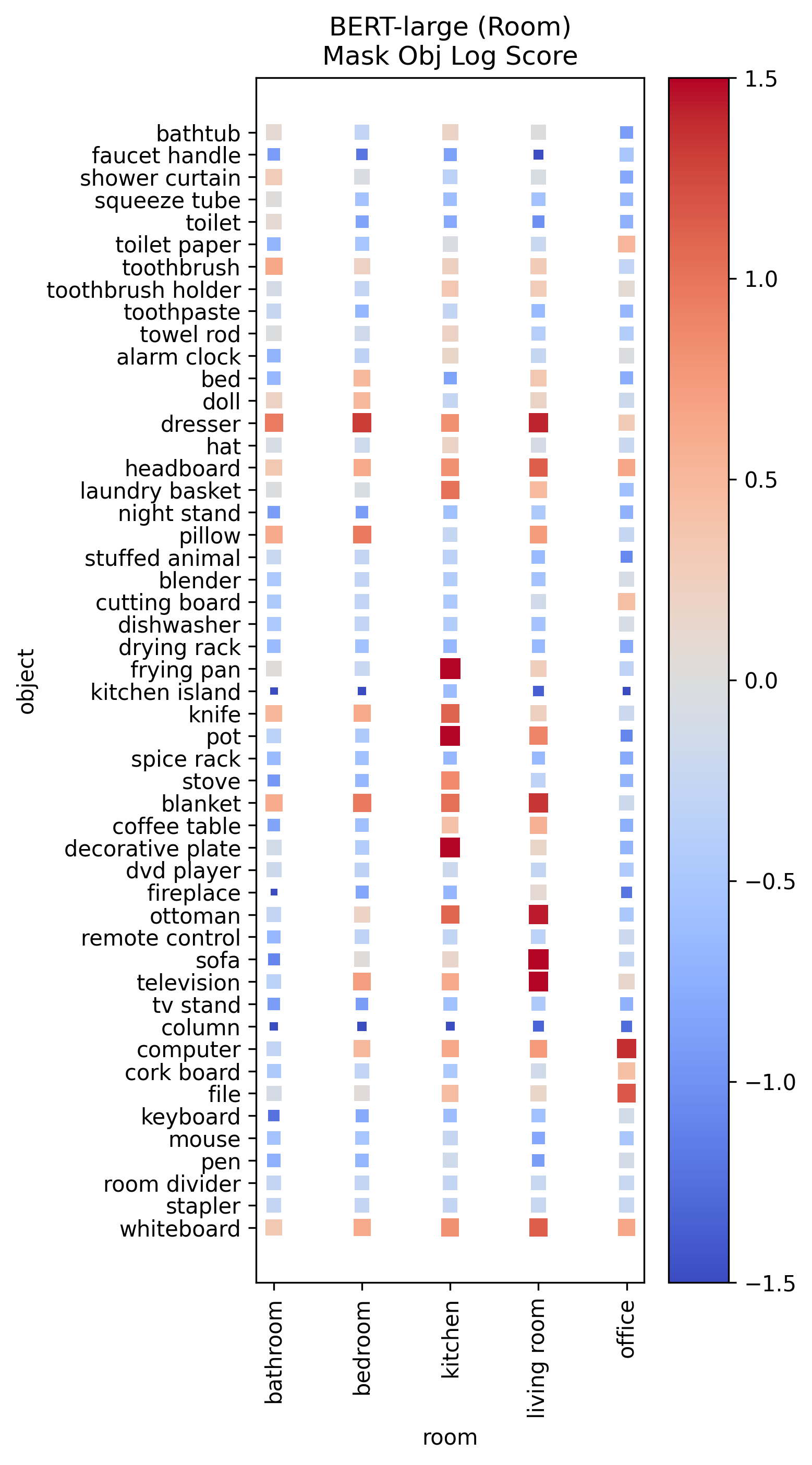}
            \vspace{-0.8cm}
            \caption{Mask Object Score}
        \end{subfigure}
        \hfill
        \begin{subfigure}[b]{0.4\textwidth}   
            \centering 
            \includegraphics[width=\textwidth, trim={0 0 0 1.2cm},clip]{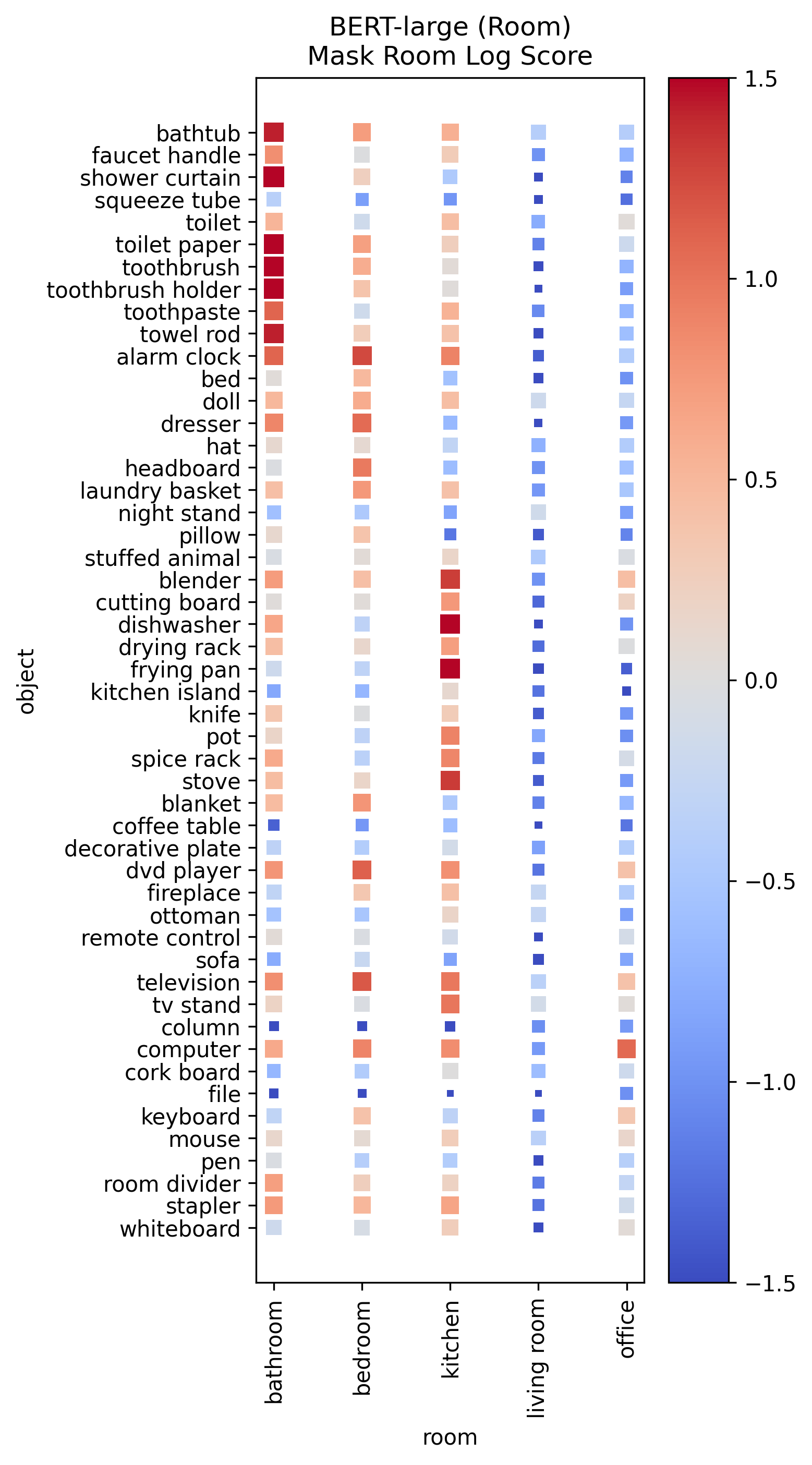}
            \vspace{-0.8cm}
            \caption{Mask Room Score}
        \end{subfigure}
        \caption{Heatmap of source-object associations based on BERT-Large and the room dataset. 
        The objects (sources) on the y-axis are grouped by the room in which they are most likely to be located according to the \textit{NYU Depth V2 Dataset}.}
        \label{fig:bert-room-all}
    \end{figure*}

\end{document}